\pdfoutput=1
\documentclass[11pt]{article}
\usepackage[T1]{fontenc}
\usepackage[utf8]{inputenc}
\usepackage{lmodern}
\usepackage[margin=1in]{geometry}
\usepackage{amsmath,amssymb}
\usepackage{graphicx}
\usepackage{float}
\usepackage{booktabs}
\usepackage{multirow}
\usepackage{xcolor}
\usepackage{tikz}
\usetikzlibrary{arrows.meta,positioning,fit,backgrounds}
\usepackage[font=small,labelfont=bf]{caption}
\usepackage{subcaption}
\usepackage[section]{placeins}

\usepackage[colorlinks=true,linkcolor={blue!70!black},citecolor={blue!70!black},urlcolor={blue!70!black}]{hyperref}
\usepackage[nameinlink]{cleveref}
\crefname{figure}{Fig.}{Figs.}\Crefname{figure}{Fig.}{Figs.}
\crefname{table}{Table}{Tables}\Crefname{table}{Table}{Tables}
\crefname{section}{Section}{Sections}

\title{\vspace{-1.8cm}Learning How a Body Moves: Few-Shot Body Inference\\for Visual World Models of Animated Characters}
\author{Amarnath R\\
{\small Department of Information Technology, United Institute of Technology, Coimbatore, India}\\
{\small\texttt{amarnathresearch@gmail.com}}}
\date{\vspace{-4ex}}

\hypersetup{pdftitle={Learning How a Body Moves: Few-Shot Body Inference for Visual World Models of Animated Characters}}
\begin{document}
\maketitle

\begin{abstract}
Two visually identical animated characters can respond differently to the same action command because their body dynamics differ. For example, a stiff blob and its bouncy twin reach different heights when both receive the jump command. Given only the first frame and the action command, a visual world model has no information to distinguish these twins, because their initial appearances are identical. We therefore propose an in-context world model that encodes clips of a character's other actions into a body representation, using a contrastive loss to separate look-alike characters. We introduce ToonTwins, a benchmark of 144 2D cartoon characters that run, jump, fall, turn and crouch. Its characters form look-alike groups of four that differ only in their dynamics profile, and we also evaluate the model on 32 animated animals from DeformingThings4D. Even with a body code from these clips, the latent pixel world model we tested recovered at most 25 percent of the twins' jump height difference. On new shapes, our model identified the correct twin in 85.9 percent of cases, compared with 76.7 percent for a transformer baseline that reads the same clips without a body summary, and in a controlled ablation the contrastive loss gave more than half of this gain. On the animated animals, it reduced the mean path prediction error by 8.5 percent against a model without body information.
\end{abstract}


\section{Introduction}

\begin{figure}[H]
\centering
\includegraphics[width=\linewidth]{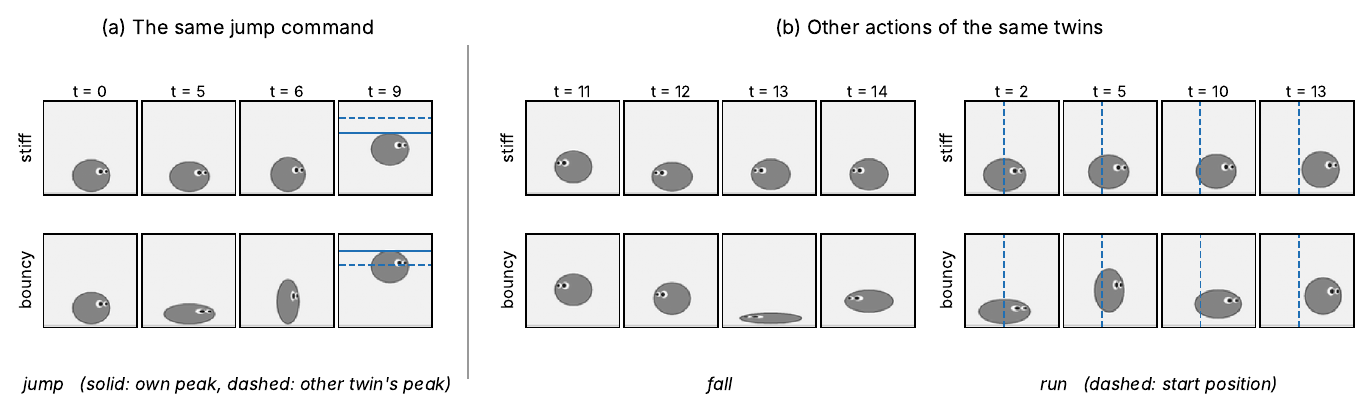}
\caption{Look-alike stiff and bouncy blobs. (a) Under the same jump command, the bouncy twin rises higher. (b) Their fall and run clips show the same difference, and the dashed line marks the start of the run.}
\label{fig:teaser}
\end{figure}

Animators make a character feel heavy or light through the way it moves rather than the way it appears. Squash and stretch, the first of the classic principles of animation \cite{thomas}, shows a soft body flattening on landing and a stiff one keeping its shape, particularly in 2D animation. This is illustrated in \hyperref[fig:teaser]{Fig.~\ref*{fig:teaser}a}, where two identical characters reach different heights in the same jump. Visual world models can generate such characters frame by frame from an action command \cite{genie,gamengen}, and others learn behavior inside a learned model of the scene \cite{dreamerv3}. However, these models predict only from the observed frames and the command, so the two characters in \hyperref[fig:teaser]{Fig.~\ref*{fig:teaser}a} provide the model with the same input, even though their true motions differ. Therefore, the input carries no information about which twin the model is predicting. The difference between their bodies becomes visible in their other actions, where the bouncy blob flattens after a fall and changes shape while running, whereas the stiff blob keeps its round shape (\hyperref[fig:teaser]{Fig.~\ref*{fig:teaser}b}). As a result, a few clips of actions such as running and falling can help a model predict how the body will jump.

Predicting embeddings instead of pixels, as in V-JEPA 2 \cite{vjepa2}, does not solve this problem, because the input still carries no evidence about the body. Standard silhouette intersection over union (IoU) also misses the error, because one predicted silhouette for both twins still overlaps each true silhouette. While prior work has inferred dynamics from recent interactions in robot control \cite{cadm,hyperdynamics} and hidden properties of look-alike objects in physical prediction \cite{physionpp,pokeworld}, in both cases these are inferred from the same video or episode in which the outcome is predicted. Inferring the body of an animated character from clips that do not contain the requested action has not been tested directly, and, to our knowledge, no video benchmark of animated characters provides look-alike pairs with different body dynamics.

In this paper, we present an in-context world model that learns how a body moves from a few clips of its other actions. This representation of the body helps predict its motion, and it does not recover physical parameters such as stiffness. The model predicts the path of the body silhouette and draws the frames by warping the first frame, so it moves the whole body and not each limb. Our contributions are as follows.
\begin{enumerate}
\item An in-context world model that predicts the motion path of a requested action from a few clips of other actions, with a contrastive loss that separates look-alike twins.
\item ToonTwins, a benchmark of 144 cartoon characters in six body plans, with 36 sets of four look-alike twins that differ only in how they move.
\item A controlled evaluation on ToonTwins and 32 animated animals from DeformingThings4D \cite{dt4d}, covering the contrastive loss, the rendered frames and unseen body plans.
\end{enumerate}

Our results show that the other actions of a character carry enough information to separate look-alike twins of new shapes, and that the contrastive loss gives more than half of the gain over a transformer baseline without a body summary. The latent pixel world model we tested loses most of this difference, even with a body code from the same clips. None of the learned models, including ours, transfers reliably to body plans absent from training.

\section{Related Work}

This section reviews visual world models, methods that infer dynamics from context, benchmarks for physical inference, motion transfer between bodies and contrastive learning.

\subsection{Visual world models}

Action-conditioned world models predict future observations from past observations and an action. Genie learns such a model from unlabeled Internet videos \cite{genie}, GameNGen simulates the game DOOM with a diffusion model \cite{gamengen}, and DreamerV3 trains a policy inside its learned model \cite{dreamerv3}. Latent-action methods learn the action space from video \cite{lapo,adaworld}, ContextWM separates the appearance of a scene from its dynamics \cite{contextwm}, and V-JEPA 2 predicts future embeddings instead of pixels \cite{vjepa2}. These models differ in their inputs and objectives, but each predicts from the observed frames and an action, so two look-alike characters give them the same input. A deterministic model trained with a pixel loss can then predict an average of the possible futures \cite{mathieu}, and stochastic latent variables can represent several possible futures and give sharper samples \cite{svg}. However, sampling a plausible future is not the same as conditioning the prediction on the correct body. In our work, the model also receives clips of the character's other actions.

\subsection{Inferring dynamics from context}

Conditional neural processes encode a few observed examples into one representation and use it to predict new outputs \cite{cnp}. Model-based reinforcement learning uses the same idea to adapt a forward model to a new environment. CaDM infers a latent vector from recent transitions \cite{cadm}, and HyperDynamics generates the parameters of the model from the agent's interactions \cite{hyperdynamics}. CoDA conditions a model of physical systems, such as ordinary and partial differential equations, on a context vector learned for each environment \cite{coda}. ICWM lets a robot policy read a few task-agnostic motions in its context window to adapt to a new camera viewpoint \cite{icwm}, and IMPULSE forecasts the V-JEPA 2 features of a pushed object from three earlier calibration pushes \cite{impulse}. These methods are the closest to ours, because each infers a system from a few observations before predicting. However, they infer a robot setup, a camera viewpoint, a physical system or a pushed object. In contrast, we infer the body dynamics of a character from actions other than the requested one, while its first frame is identical to that of its look-alike twins.

\subsection{Benchmarks for physical inference}

Physics-IQ shows that video generation models have a weak understanding of physics, and that this understanding is unrelated to visual realism \cite{physicsiq}. Physion++ shows the property of an object during an inference period and then asks for a prediction from a scene that looks the same for every property value, so the outcome can only be predicted by inferring mass, friction, elasticity or deformability from motion \cite{physionpp}. PokeWorld pushes visually identical objects and tests which hidden properties are encoded in the latent states of a world model \cite{pokeworld}. PAC-NeRF and PhysDreamer recover material parameters of 3D objects with a physics simulator \cite{pacnerf,physdreamer}. ToonTwins follows Physion++ and PokeWorld in keeping the appearance identical while the dynamics differ. However, our characters perform their own actions instead of being pushed, and our model needs no simulator.

\subsection{Motion transfer between bodies}

Motion retargeting maps a given motion onto another skeleton. Prior methods handle skeletons with homeomorphic structures \cite{aberman}, with different numbers of joints \cite{palum} and with arbitrary topologies \cite{topocap}. Image animation moves a character image with keypoints \cite{fomm}, a pose sequence \cite{animateanyone} or drag trajectories \cite{puppetmaster}. In these methods, the motion is given as input and is transferred to another body or image. In our task, the requested action has not been observed for the character, and the model has to predict it from the character's other actions.

\subsection{Contrastive learning}

The InfoNCE loss trains an encoder to pick the positive example for each anchor from a set of negatives \cite{cpc}, so what the representation keeps depends on how the pairs are built. In our model, the positive pairs are summaries of one character computed from two disjoint sets of clips, and the negatives include its look-alike twins. This trains the summary to keep the information that identifies the body across different actions. The loss does not remove all action information, so we measure what remains with linear probes in \cref{sec:invariance}.

\subsection{Summary}

None of the methods above predicts the motion of a character in video from clips of actions other than the requested one when appearance cannot reveal the body. \Cref{sec:problem} defines the task formally.

\section{Benchmark and Datasets}
\label{sec:data}

To test whether the other actions of a character reveal how its body moves, the data must meet three requirements.
\begin{enumerate}
\item The characters must look alike, but their movements must differ.
\item Each character must perform several labeled actions.
\item Each clip must come with the action command that produced it.
\end{enumerate}
To our knowledge, no existing dataset in the literature meets all three requirements. However, several datasets meet some of them. Sprites \cite{sprites} has labeled actions but fails the first requirement, because all its characters share one human body and one set of animations, so they differ in appearance but not in motion. General videos of people or animals fail the first and third requirements, because a change in appearance usually comes with a change in the body and the videos have no action commands. Physion++ \cite{physionpp} and PokeWorld \cite{pokeworld} meet the first requirement for objects, but they fail the second, because their objects are pushed or collide instead of performing labeled actions. We therefore built ToonTwins, in which look-alike twins differ only in their dynamics profile (\cref{sec:toontwins}), and we also use animated animals from DeformingThings4D \cite{dt4d} to check whether the findings hold for more complex bodies (\cref{sec:animals}).

\subsection{ToonTwins}
\label{sec:toontwins}

ToonTwins contains 144 cartoon characters, which we built from 6 body plans (biped, blob, disjointed, limbless, multilimb and quadruped), 3 shapes, 2 appearances and 4 dynamics profiles. The three shapes change the proportions of the body. The first keeps the default width and height, the second makes the body 16 percent wider and 12 percent shorter, and the third makes it 12 percent narrower and 12 percent taller. The two appearances use different colors, patterns (plain or spotted) and eye colors. Characters with the same body plan, shape and appearance look identical, so the benchmark has 36 sets of four look-alike twins.

\begin{figure}[H]
\centering
\begin{subfigure}{\linewidth}\centering\includegraphics[width=0.78\linewidth]{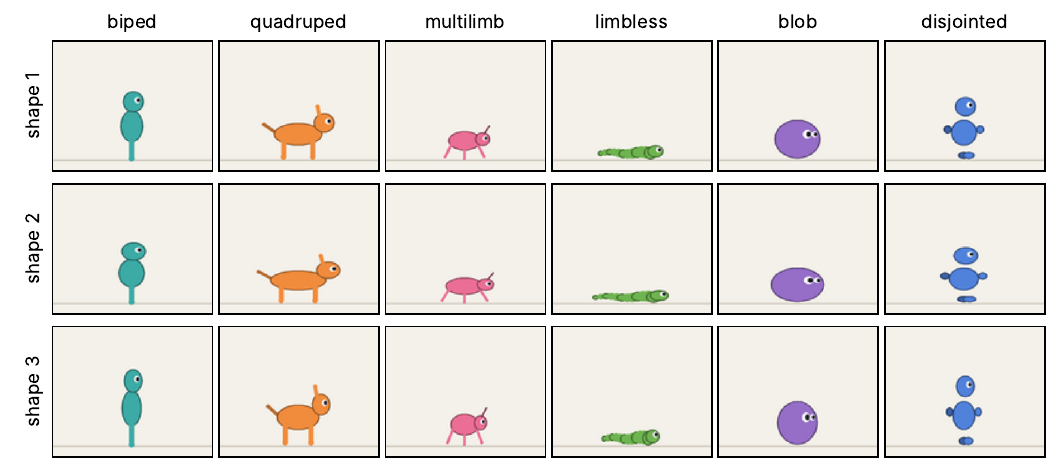}\caption{The 6 body plans in their 3 shapes (first frames). Each image stands for four look-alike twins.}\end{subfigure}\\[4pt]
\begin{subfigure}{\linewidth}\centering\includegraphics[width=0.72\linewidth]{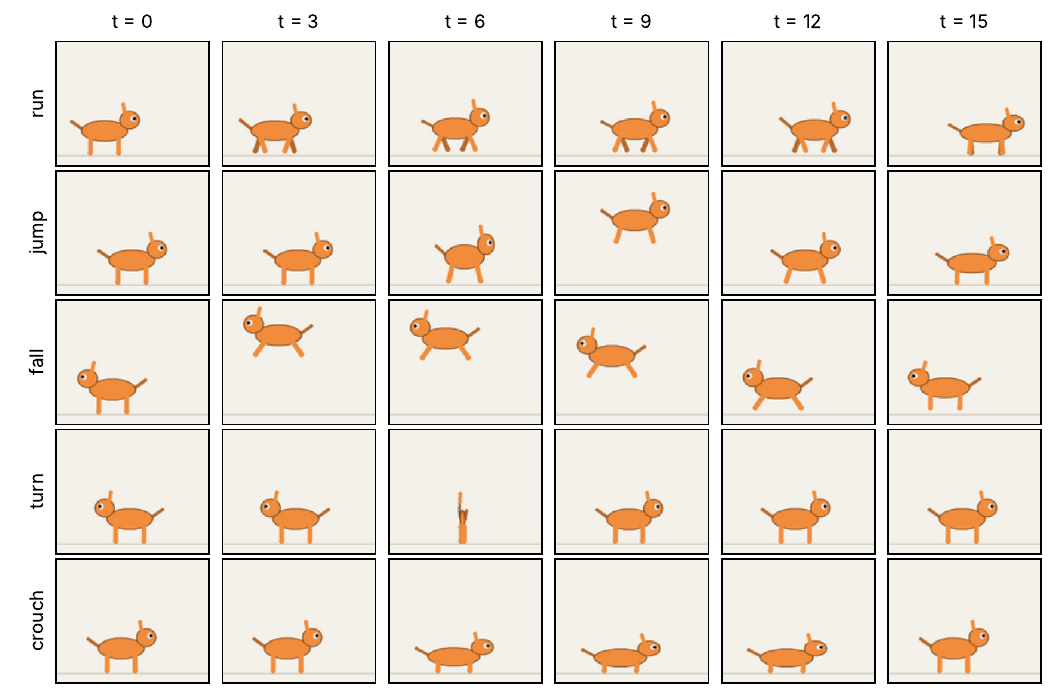}\caption{The 5 actions of one character (bouncy quadruped).}\end{subfigure}\\[4pt]
\begin{subfigure}{\linewidth}\centering\includegraphics[width=0.7\linewidth]{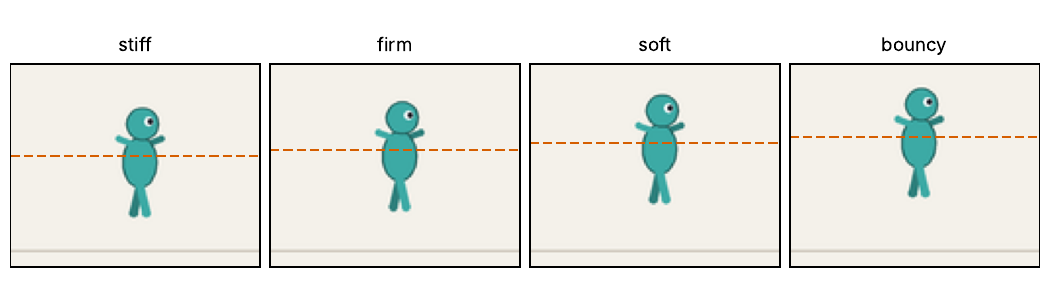}\caption{Four look-alike twins (biped) at the peak of the same jump. The dashed line marks the body centroid.}\end{subfigure}
\caption{ToonTwins. Characters are shown in color for display, and all models receive the same frames in gray.}
\label{fig:bench}
\end{figure}

The four twins in a set differ only in their dynamics profile, which is stiff, firm, soft or bouncy. Each body plan has its own default values for eight dynamics parameters, which control the jump height, the bounce and squash on landing, the body bob, the gait, the running speed, the crouch depth and the fall speed, and the profile multiplies these defaults by fixed factors. For example, the bouncy profile raises the jump by 22 percent and the rebound after a fall by 130 percent over the default of the body plan, while the stiff profile lowers them by 22 and 65 percent (\cref{fig:teaser} and \hyperref[fig:bench]{Fig.~\ref*{fig:bench}c}). These multipliers are fixed design choices rather than values measured from data. We chose them so that every parameter changes monotonically from stiff to bouncy. Squash and bounce, which define squash and stretch in animation \cite{thomas}, have the widest range. In contrast, the running speed changes by at most 8 percent, which keeps every character inside the $64\times 64$ frame and prevents position alone from identifying the twin. \hyperref[app:params]{Appendix~\ref*{app:params}} lists all default values and multipliers.

Each character performs 5 actions (run, jump, fall, turn and crouch) under 12 intents, where an intent sets the effort, the direction, the start position and the number of standing frames (1 to 4) before the action starts. This gives 8,640 clips of 16 gray frames at $64\times 64$ pixels (\cref{fig:bench}).

This design gives the benchmark four properties. The standing frames do not depend on the dynamics profile, so the four twins of a set have identical first frames in every clip. Their motion, however, differs, and at the peak of a jump the silhouette centroid rises by 7.57\,px on average for the stiff twin and by 12.16\,px for the bouncy twin. Turning is the only action that does not depend on the profile, so we use it as a control action. Finally, a linear probe \cite{probes} on simple statistics of the silhouette paths identifies the profile from the other actions of a character in 0.506 of cases, against a chance level of 0.25 (\cref{sec:invariance}).

\subsection{Animated animals}
\label{sec:animals}

\begin{figure}[!htb]
\centering
\begin{subfigure}{\linewidth}\centering\includegraphics[width=0.75\linewidth]{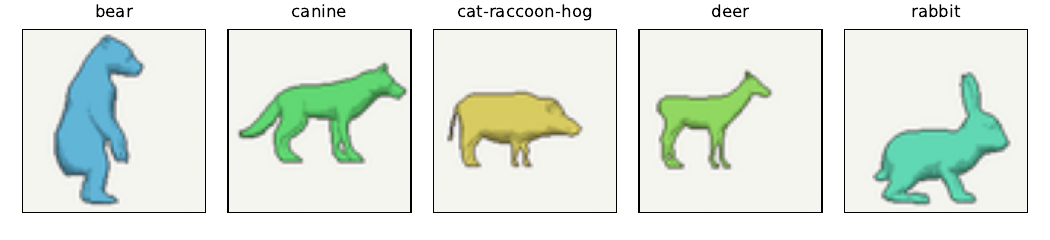}\caption{One animal from each family.}\end{subfigure}\\[4pt]
\begin{subfigure}{\linewidth}\centering\includegraphics[width=0.36\linewidth]{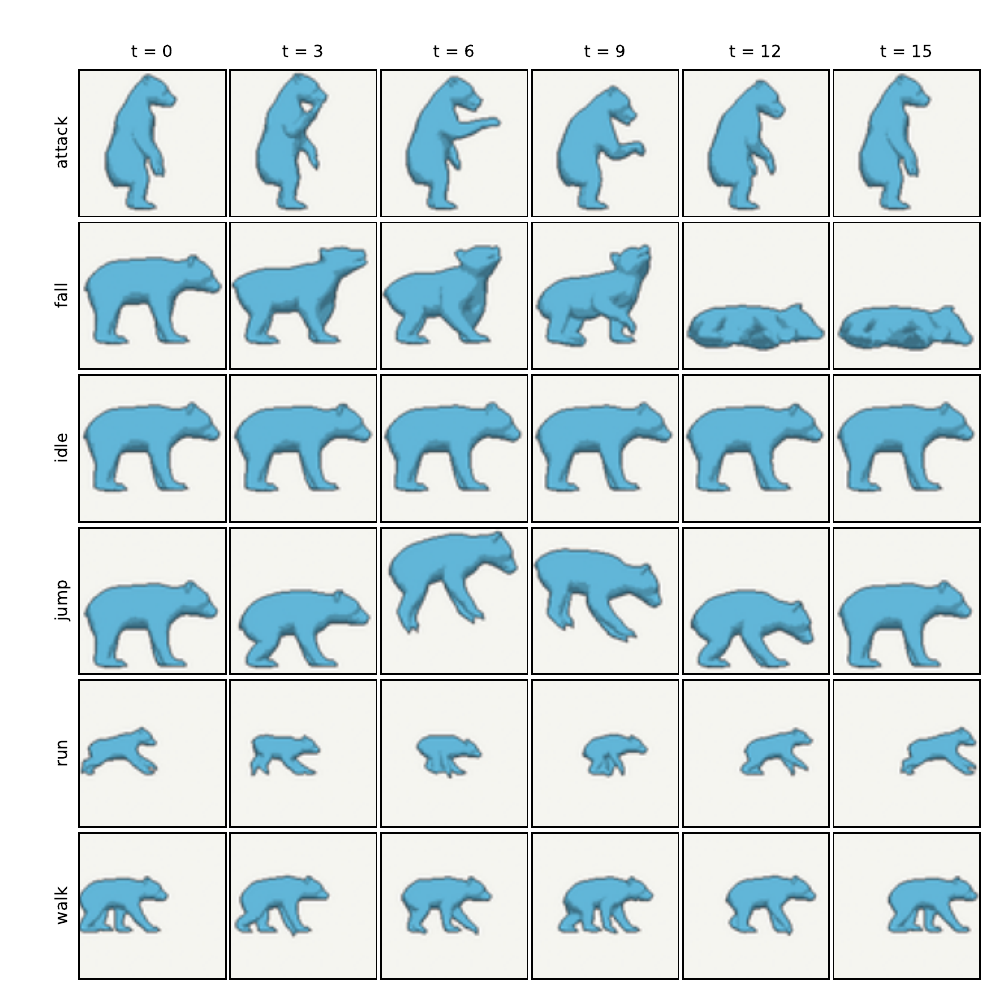}\caption{One animal (bear) performing its six actions.}\end{subfigure}
\caption{Animated animals from DeformingThings4D \cite{dt4d}.}
\label{fig:realdata}
\end{figure}

DeformingThings4D contains 3D animations of humanoids and animals made by artists \cite{dt4d}, and each animation shows one character performing one action. From its animal animations, we kept the animals that perform at least four of six common actions (attack, fall, idle, jump, run and walk), which gives 169 animations of 32 animals from 5 families (bear, canine, \mbox{cat-raccoon-hog}, deer and rabbit). We render each animation as a flat 2D side view with three shading tones and a dark outline on a light background (\cref{fig:realdata}). The camera looks across the long horizontal axis of the body in the first frame, and each animal is mirrored to face right. A clip holds 16 frames spaced evenly over the animation, drawn at $128\times128$ pixels, reduced to $64\times64$ and converted to gray. The camera follows the center of the body box in each frame with a fixed scale for the whole animation, so the clips keep the changes of pose and shape but not the travel across the scene. For training only, we also render clips from the first 75 percent and the last 75 percent of each animation. Some animals in the dataset are copies of another animal with a different skin but the same motion. To find them, we compared every pair of animals on each action they share. We describe an animation by six curves over time, the three coordinates of the mesh centroid and the three side lengths of the bounding box, and compare two animations of equal length by the Pearson correlation between the frame-to-frame changes of each curve, averaged over the six curves. Two animals share their animations when at least one animation is identical on the same mesh, with a relative difference in vertex motion below 0.05, or when at least two animations of equal length, covering at least half of the shared actions, have a mean correlation of at least 0.90. We merged the groups identified by two runs of this procedure, one of which used an earlier version of the comparison method. This resulted in 21 independent motion groups. A first grouping missed four animals that share animations and gave 24 groups, and all animal results in this paper use the corrected 21 groups, except the pixel world model test in \cref{sec:real}. Each animal belongs to one group, and each group stays on one side of every split, so a test animal and all its actions are absent from training. Because these animals have no twins, appearance and dynamics remain mixed in this data, unlike in ToonTwins.

\section{Method}

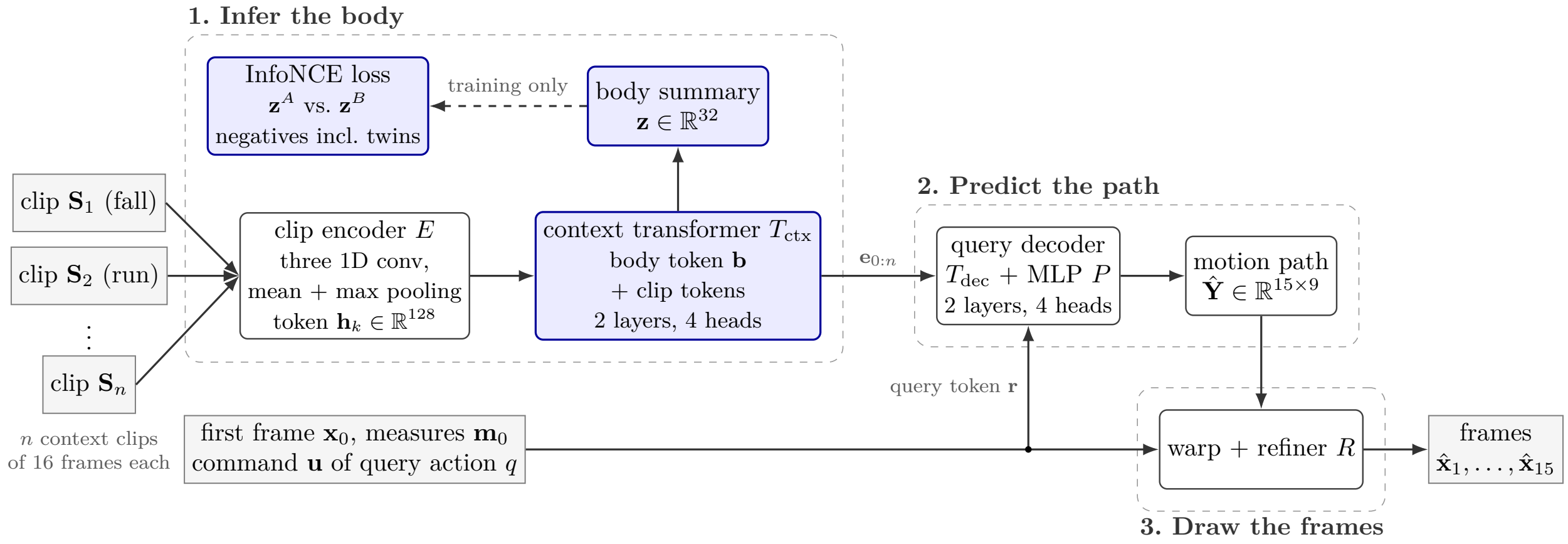
\begin{figure}[H]
\centering
\resizebox{\linewidth}{!}{%
\begin{tikzpicture}[
  font=\small, node distance=6mm and 9mm,
  box/.style={draw=black!75, rounded corners=3pt, align=center, minimum height=11mm, inner sep=3pt, fill=white, line width=0.6pt},
  ours/.style={draw=blue!60!black, rounded corners=3pt, align=center, minimum height=11mm, inner sep=3pt, fill=blue!8, line width=0.8pt},
  data/.style={draw=black!50, align=center, minimum height=8mm, inner sep=3pt, fill=black!4, line width=0.5pt},
  arr/.style={-{Latex[length=2.2mm]}, line width=0.8pt, black!80},
  dim/.style={font=\scriptsize, text=black!65},
  stage/.style={draw=black!40, dashed, rounded corners=6pt, inner sep=3mm},
  lab/.style={font=\small\bfseries, text=black!80, anchor=south west, inner sep=1pt, yshift=1pt}]
\node[data] (c1) {clip $\mathbf{S}_1$ (fall)};
\node[data, below=2mm of c1] (c2) {clip $\mathbf{S}_2$ (run)};
\node[below=-0.5mm of c2] (dots) {$\vdots$};
\node[data, below=-0.5mm of dots] (cn) {clip $\mathbf{S}_n$};
\node[dim, below=1mm of cn, align=center] {$n$ context clips\\of 16 frames each};
\node[box, right=10mm of c2] (ce) {clip encoder $E$\\{\footnotesize three 1D conv,}\\{\footnotesize mean + max pooling}\\{\footnotesize token $\mathbf{h}_k\in\mathbb{R}^{128}$}};
\node[ours, right=of ce] (te) {context transformer $T_{\mathrm{ctx}}$\\{\footnotesize body token $\mathbf{b}$}\\{\footnotesize + clip tokens}\\{\footnotesize 2 layers, 4 heads}};
\node[ours, above=9mm of te] (z) {body summary\\$\mathbf{z}\in\mathbb{R}^{32}$};
\node[ours, left=22mm of z] (nce) {InfoNCE loss\\{\footnotesize $\mathbf{z}^{A}$ vs.\ $\mathbf{z}^{B}$}\\{\footnotesize negatives incl.\ twins}};
\node[box, right=16mm of te] (de) {query decoder\\$T_{\mathrm{dec}}$ + MLP $P$\\{\footnotesize 2 layers, 4 heads}};
\node[box, right=of de] (path) {motion path\\$\hat{\mathbf{Y}}\in\mathbb{R}^{15\times 9}$};
\node[box, below=13mm of path] (rend) {warp + refiner $R$};
\node[data, right=of rend] (fr) {frames\\$\hat{\mathbf{x}}_1,\dots,\hat{\mathbf{x}}_{15}$};
\node[data] (q) at (ce |- rend) {first frame $\mathbf{x}_0$, measures $\mathbf{m}_0$\\command $\mathbf{u}$ of query action $q$};
\begin{scope}[on background layer]
\node[stage, fit=(ce)(te)(z)(nce)] (s1) {};
\node[stage, fit=(de)(path)] (s2) {};
\node[stage, fit=(rend)] (s3) {};
\end{scope}
\node[lab] at (s1.north west) {1.\ Infer the body};
\node[lab] at (s2.north west) {2.\ Predict the path};
\node[lab, anchor=north west, yshift=-2pt] at (s3.south west) {3.\ Draw the frames};
\foreach \c in {c1,c2,cn} \draw[arr] (\c.east) -- (ce.west);
\draw[arr] (ce) -- (te);
\draw[arr] (te) -- (z);
\draw[arr, dashed] (z) -- node[above, dim]{training only} (nce);
\draw[arr] (te) -- node[above, dim]{$\mathbf{e}_{0:n}$} (de);
\draw[arr] (de) -- (path);
\draw[arr] (path) -- (rend);
\draw[arr] (rend) -- (fr);
\draw[arr] (q.east) -- (rend.west);
\draw[arr] (q.east -| de) -- node[left, dim]{query token $\mathbf{r}$} (de.south);
\fill (q.east -| de) circle (1.3pt);
\end{tikzpicture}}
\caption{Overview of the model. A clip encoder and a context transformer summarize the context clips, a query decoder predicts the path and a renderer draws the frames.}
\label{fig:method}
\end{figure}

Our model works in three stages (\cref{fig:method}). It encodes the context clips into tokens, predicts the motion path of the requested action from these tokens and renders the frames along this path.

\subsection{Problem formulation}
\label{sec:problem}

Our task is to predict how a character performs a requested action, which we call the query action $q$, from three inputs. The first input is the first frame $\mathbf{x}_0$ of the query clip. The second is the command $\mathbf{u}$ of the query action. On ToonTwins, $\mathbf{u}\in\mathbb{R}^{10}$ holds a one-hot code of the five action types, the effort and a one-hot code of the four onset frames, and on the animated animals, $\mathbf{u}\in\mathbb{R}^{6}$ is a one-hot code of the six actions. The third is the context set $\mathcal{C}=\{\mathbf{S}_1,\dots,\mathbf{S}_n\}$, which holds $n$ clips of the same character performing actions other than $q$, and each clip $\mathbf{S}_k$ ($k=1,\dots,n$) covers 16 frames.

\begin{figure}[H]
\centering
\includegraphics[width=0.75\linewidth]{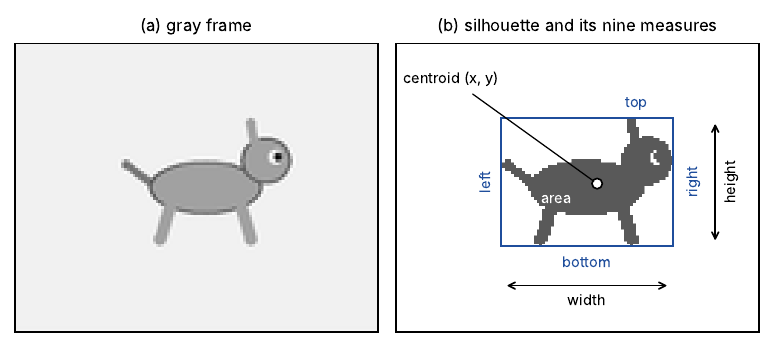}
\caption{Input of the path model. (a) A gray frame. (b) Its silhouette with the nine measures, which are the box edges, the area, the centroid, the height and the width.}
\label{fig:silhouette}
\end{figure}

We describe the body by its silhouette, which we obtain by thresholding each frame against the background and removing the ground strip. From each silhouette, we take nine measures, which are the four edges of the bounding box, the area, the two centroid coordinates, the height and the width, and we write $\mathbf{m}_0\in\mathbb{R}^9$ for the measures of the first frame (\cref{fig:silhouette}). The model $f_\theta$, with learnable parameters $\theta$, predicts the silhouette path of the next 15 frames,
\begin{equation}
\hat{\mathbf{Y}}=f_\theta\left(\mathbf{m}_0,\mathbf{u},\mathcal{C}\right)\in\mathbb{R}^{15\times 9},
\label{eq:task}
\end{equation}
where row $t$ of $\hat{\mathbf{Y}}$ holds the change of the nine measures from the first frame to frame $t$, and $\mathbf{Y}$ denotes the true path. The frames $\hat{\mathbf{x}}_1,\dots,\hat{\mathbf{x}}_{15}$ are then drawn from $\hat{\mathbf{Y}}$ and $\mathbf{x}_0$ (\cref{sec:render}). Since $q$ is never in $\mathcal{C}$, the model can infer how the character performs $q$ only from its other actions. Look-alike twins also share $\mathbf{x}_0$, $\mathbf{m}_0$ and $\mathbf{u}$, so only $\mathcal{C}$ can make their predictions differ.

\subsection{Clip encoder}

The clip encoder turns each context clip into one vector, which we call a clip token. For every frame of the clip, the input holds the nine measures relative to the first frame of the clip, their change from the previous frame and the command of the clip, so a clip is a matrix $\mathbf{S}_k\in\mathbb{R}^{16\times 28}$ on ToonTwins and $\mathbf{S}_k\in\mathbb{R}^{16\times 24}$ on the animated animals. Three 1D convolution layers $\phi$ process this matrix over time, and the mean and the maximum over time give the clip token
\begin{equation}
\mathbf{h}_k=W_h\left[\operatorname*{mean}_t \phi(\mathbf{S}_k)\,;\,\max_t \phi(\mathbf{S}_k)\right]\in\mathbb{R}^{128},
\label{eq:clip}
\end{equation}
where $[\cdot\,;\cdot]$ joins the two pooled vectors of 128 values each and $W_h\in\mathbb{R}^{128\times 256}$ is a learned matrix.

\subsection{In-context transformer with a body summary}

Our model keeps the clip tokens separate and combines them with a transformer in two steps. First, a context transformer $T_{\mathrm{ctx}}$ \cite{vaswani} reads the clip tokens together with an extra learnable token $\mathbf{b}$, which we call the body token,
\begin{equation}
\left[\mathbf{e}_0,\mathbf{e}_1,\dots,\mathbf{e}_n\right]=T_{\mathrm{ctx}}\left(\left[\mathbf{b}+\boldsymbol{\eta}_{b},\;\mathbf{h}_1+\boldsymbol{\eta}_{c},\;\dots,\;\mathbf{h}_n+\boldsymbol{\eta}_{c}\right]\right),\qquad \mathbf{z}=g(\mathbf{e}_0)\in\mathbb{R}^{32},
\label{eq:ctx}
\end{equation}
where $\boldsymbol{\eta}_b$ and $\boldsymbol{\eta}_c$ are learned embeddings that mark the two token types. The body token gathers information about the character from all clips, and its output $\mathbf{e}_0$, passed through a small network $g$, gives the body summary $\mathbf{z}$. Since $T_{\mathrm{ctx}}$ never sees the query action, the summary cannot depend on it.

Second, a query decoder $T_{\mathrm{dec}}$ reads a query token $\mathbf{r}=W_r[\mathbf{m}_0;\mathbf{u}]+\boldsymbol{\eta}_q$ together with all outputs of the context transformer, and a predictor network $P$ turns the decoder output at the query token into the path,
\begin{equation}
\hat{\mathbf{Y}}=P\left(T_{\mathrm{dec}}\left(\left[\mathbf{r},\mathbf{e}_0,\mathbf{e}_1,\dots,\mathbf{e}_n\right]\right)_{\mathbf{r}}\right),
\label{eq:dec}
\end{equation}
where the subscript $\mathbf{r}$ selects the output at the query token. The decoder reads both $\mathbf{e}_0$ and the clip outputs $\mathbf{e}_1,\dots,\mathbf{e}_n$. The summary $\mathbf{z}$ is not used for prediction. It is used only by the contrastive loss (\cref{sec:contrastive}) and by the probes (\cref{sec:invariance}), and through it the contrastive loss also trains $\mathbf{e}_0$. We train the path with the mean squared error on standardized values,
\begin{equation}
\mathcal{L}_{\mathrm{pred}}=\frac{1}{15\cdot 9}\left\lVert \hat{\mathbf{Y}}-\mathbf{Y}\right\rVert_F^2 .
\label{eq:pred}
\end{equation}

\subsection{Contrastive objective for a body-specific summary}
\label{sec:contrastive}

With the prediction loss alone, the summary can also store which actions appear in the context, because this information lowers the prediction error. We want the summary to describe the body rather than the observed actions, so we add a contrastive loss that makes the summaries of one character agree for different observed actions. The loss reduces this action information but does not remove it, which we measure in \cref{sec:invariance}.

For the $i$-th character in a batch of $N$ characters, we draw two context sets $\mathcal{C}^{A}_i$ and $\mathcal{C}^{B}_i$ that share no clip and compute their summaries $\mathbf{z}^{A}_i$ and $\mathbf{z}^{B}_i$. The two summaries of the same character form a positive pair, and the summaries of the other characters are the negatives,
\begin{equation}
\mathcal{L}_{c} = -\frac{1}{2N}\sum_{i=1}^{N}\left[\log\frac{e^{\langle \hat{\mathbf{z}}^{A}_{i}, \hat{\mathbf{z}}^{B}_{i}\rangle/\tau}}{\sum_{j}e^{\langle \hat{\mathbf{z}}^{A}_{i}, \hat{\mathbf{z}}^{B}_{j}\rangle/\tau}} + \log\frac{e^{\langle \hat{\mathbf{z}}^{B}_{i}, \hat{\mathbf{z}}^{A}_{i}\rangle/\tau}}{\sum_{j}e^{\langle \hat{\mathbf{z}}^{B}_{i}, \hat{\mathbf{z}}^{A}_{j}\rangle/\tau}}\right],
\label{eq:nce}
\end{equation}
where $\hat{\mathbf{z}}$ is $\mathbf{z}$ scaled to unit length and $\tau=0.1$ is a temperature. Each batch holds 16 complete twin sets, so the hardest negatives of a character are its own look-alike twins, which look the same but move differently. The other color version of the character moves identically, so we remove it from both sums, and on the animated animals we remove the animals of the same motion group. The total loss is
\begin{equation}
\mathcal{L}=\mathcal{L}_{\mathrm{pred}}+\lambda\,\mathcal{L}_{c},\qquad \lambda=0.5 .
\label{eq:total}
\end{equation}
We also train a stricter variant in which the two context sets share no action type (two action types each on ToonTwins), which tests whether the summary stays the same when the observed actions differ.

\subsection{Path-guided rendering}
\label{sec:render}

The renderer draws each future frame by moving and stretching the first frame so that its bounding box matches the predicted box. For frame $t$, a scale and shift $A_t$ maps the box in $\mathbf{x}_0$ to the predicted box, which is given by $\mathbf{m}_0+\hat{\mathbf{y}}_t$, and the warp $\mathcal{W}$ resamples the frame with bilinear interpolation while the ground line stays fixed. A refinement network $R$ then corrects local details such as the positions of the limbs,
\begin{equation}
\mathbf{w}_t=\mathcal{W}\left(\mathbf{x}_0;A_t\right),\qquad \hat{\mathbf{x}}_t=\mathbf{w}_t+R\left(\mathbf{w}_t,\mathbf{x}_0,\hat{\mathbf{y}}_t\right).
\label{eq:render}
\end{equation}
The network $R$ is trained only on the true paths of training characters, with small noise on the box edges, and never sees the output of a path model, so we use the same renderer for every model we compare.

\subsection{Training details}

The convolution layers of the clip encoder have 128 channels, kernel size 3 and GELU activations \cite{gelu}. Both transformers have two pre-norm layers \cite{prenorm} of width 128, four attention heads, a feed-forward size of 512 and no dropout. The network $g$ is a two-layer MLP, and the predictor networks have three hidden layers of 512 units. The refinement network $R$ is a three-level U-Net \cite{unet} with 32, 64 and 128 channels, trained for 3,000 steps with 0.7\,px of noise on the box edges and an L1 plus binary cross-entropy loss.

All path models use AdamW \cite{adamw} with a learning rate of $10^{-3}$, a one-cycle schedule \cite{onecycle}, gradient clipping at 1.0 and a batch of 64 characters. On ToonTwins, we train for 3,000 steps with a weight decay of $10^{-4}$ and draw the context size $n$ uniformly from 1 to 8 at every step. On the animated animals, we train for 2,000 steps with a weight decay of $10^{-3}$ and draw $n$ from 1 to 3, because some animals perform only four actions, which leaves three for the context. Training one model takes 11 to 64\,s on a single Tesla T4 GPU.

\section{Experimental Setup}

\subsection{Splits}

We use two splits on ToonTwins. In the split with new shapes, each of 3 folds holds out one shape of every body plan, and we train with 3 seeds. In the split with unseen body plans, each of 6 folds holds out one whole body plan, and we train with 2 seeds. On the animated animals, we use 4 folds over the motion groups and 3 seeds. In all splits, the context of a test character holds only clips of actions other than the query action.

\subsection{Models compared}

\Cref{tab:models} lists the models we compare, which differ mainly in the body information given to each model. All path models use the same clip encoder and predictor network. The body code averages the clip tokens of the context set and maps the average to a 32-dimensional vector with a two-layer MLP, and the predictor maps $\mathbf{m}_0$, $\mathbf{u}$ and this vector to the path. The context attention baseline is a single transformer with three layers that reads the query token and the clip tokens together and predicts the path from the output at the query token. It has no body token and no contrastive loss, and its clip tokens are processed together with the query. Our model adds the body token, the contrastive loss and a context transformer that does not see the query. The profile or family label is a reference and not an upper bound, because the context clips carry more than the label, for example how a character of this exact shape moves.

\begin{table}[H]
\centering
\caption{Models compared.}
\label{tab:models}
\small
\begin{tabular}{p{0.34\linewidth}p{0.46\linewidth}r}
\toprule
Model & Body information it receives & Parameters \\
\midrule
No body & First frame and action command only & 0.60\,M \\
Fine-tune at test & No-body model fine-tuned for 50 steps on the context clips & 0.60\,M \\
Body code & Single code from the context clips & 0.78\,M \\
Body code + contrastive loss & Same, with the contrastive loss & 0.78\,M \\
Context attention & Query attends over the context clip tokens & 1.40\,M \\
\textbf{Ours} & Body token, query-free context encoder and contrastive loss & 1.62\,M \\
Profile or family label & True dynamics profile (ToonTwins) or animal family (animals), as a reference & 0.61\,M \\
Pixel world model & Latent world model decoding frames, conditioned on a body code & 2.88\,M \\
Copy first frame & Repeats the first frame & -- \\
\bottomrule
\end{tabular}
\end{table}

The pixel world model predicts frames instead of paths. It encodes each $48\times 48$ frame into a 128-dimensional embedding with three convolution layers. A residual MLP with 512 units then predicts the next embedding, and it receives a 32-dimensional body code and the full action command through FiLM layers \cite{film}. Three transposed convolutions decode the embedding back into a frame. The body code comes from the same kind of context set as for the path models, with or without the contrastive loss. We train this model for 3,000 steps with batches of 16 clips of 12 frames. The loss has four parts, which are a one-step loss on the embeddings, an open-loop rollout loss from the first frame, a regularizer on the embeddings and a decoder loss on detached embeddings. Training takes 150 to 200\,s on a Tesla T4. The rollout loss was still falling slowly at the end, so longer training could improve this model. Our conclusions therefore hold only for this model and not for pixel world models in general.

\subsection{Measures}

On ToonTwins, our main measure is \emph{twin matching}. We first remove what the four twins of a set share from both the predicted and the true paths. We then assign each predicted twin to the nearest true twin and report the fraction of correct assignments, where chance is 0.25. The \emph{jump gain} measures how much of the height difference between twins is recovered. It is the slope of the predicted twin differences in peak height against the true differences, so a value of 1 means full recovery of the difference and a value of 0 means no recovery. We also report the Spearman correlation \cite{spearman} of the four peak heights, how often the highest jumper is found (chance 0.25) and a swap test in which each twin receives the context of another twin.

On the animated animals, we report the path error in standardized units and the error reduction
\begin{equation}
R=100\,\frac{\bar{E}_{\mathrm{none}}-\bar{E}}{\bar{E}_{\mathrm{none}}},
\label{eq:reduction}
\end{equation}
where $\bar{E}$ is the mean error of a model and $\bar{E}_{\mathrm{none}}$ that of the model without body information. To compute each mean, we average the error over the queries, folds and seeds within each motion group and then over the motion groups, so each motion group has equal weight. We first planned to compute this percentage for each query and average it, but a few queries with a very small error of the no-body model give extreme values, as low as $-401$ percent, so we report that version only as a check. We also report \emph{context identification}, which checks whether an animal's own context gives a lower error than the context of each of up to 7 held-out animals from other motion groups (chance 0.157).

We compare rendered frames with the truth at $48\times 48$ pixels. For them, we report the IoU at a threshold of 0.25, a sharpness ratio and twin matching computed on the images. The sharpness ratio divides the image gradient energy of the prediction by that of the truth, so 1 means as sharp as the truth.

\subsection{Statistics}

On ToonTwins, the unit of analysis is a pair of body plan and shape, which gives 18 units. On the animated animals, it is a motion group, which gives 21 units. We first average the folds and seeds within each unit. We then report means with 95 percent bootstrap confidence intervals \cite{efron} over 2,000 resamples of units. To compare two models, we use paired bootstrap and Wilcoxon signed-rank tests \cite{wilcoxon} over units and report Cohen's $d_z$ \cite{cohen} as the paired effect size. The representation analysis in \cref{sec:invariance} uses its own procedure, which we describe in that section, because its runs share test characters.

\section{Results}

We first test whether the pixel world model keeps the twin difference and then whether path prediction from the context recovers it. We then separate the parts of our model, test its body summary and its rendered frames, and repeat the main test on the animated animals and on unseen body plans.

\subsection{The tested pixel world model loses body-specific dynamics}

The pixel world model we tested has more parameters than our model (2.88\,M against 1.62\,M), but its decoded rollouts match the twins in only 0.27 to 0.30 of cases, where chance is 0.25, and recover only 14 to 25 percent of the true difference in jump height. Its IoU of 0.44 to 0.45 is lower than that of copying the first frame (0.50), and its frames are blurred, with a sharpness of 0.66 relative to the truth. This failure does not depend on the form of the command, because a variant that receives only the action type, without the effort or onset, matched the twins in 0.29 to 0.31 of cases.

\begin{figure}[H]
\centering
\includegraphics[width=\linewidth]{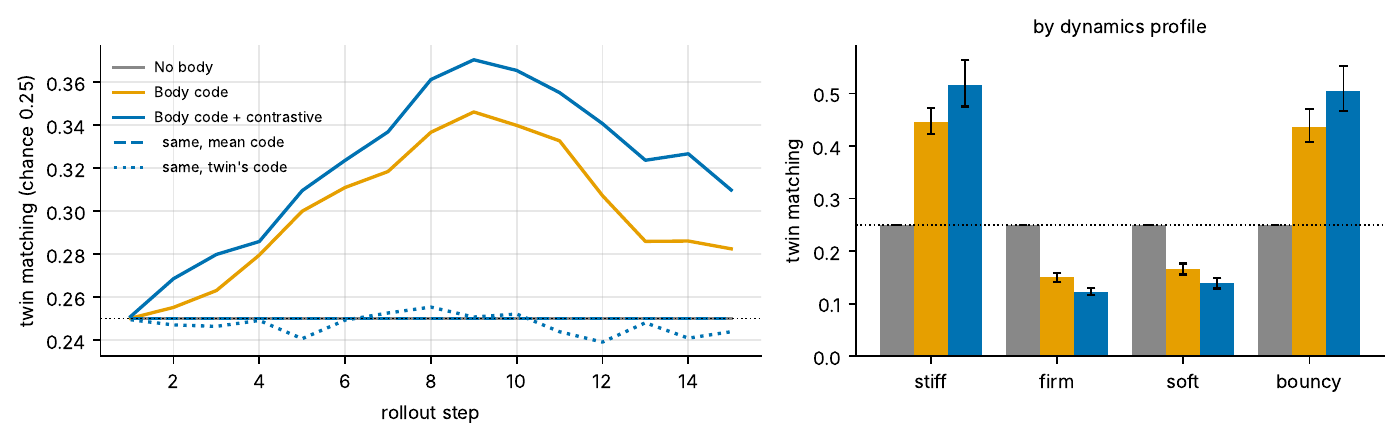}
\caption{Twin matching of the pixel world model on new shapes (turn excluded), over the rollout steps (left) and by dynamics profile (right), with 95\% CIs over units.}
\label{fig:exp3pixel}
\end{figure}

\Cref{fig:exp3pixel} shows where the small amount of body information appears in the rollouts. Twin matching rises only during the flight phase of the jump and the run, peaks at 0.37 at step 9 with the contrastive body code and falls again afterwards. Replacing the code with the mean code or with a twin's code brings it back to chance, so this small effect does come from the body code. The model also separates only the extreme profiles. Stiff and bouncy twins are matched in 0.52 and 0.51 of cases, but firm and soft twins in only 0.12 and 0.14, which is below chance. This is because the differences between the four predictions are small compared with the true differences, and only the extremes are sometimes ordered correctly.

\begin{figure}[H]
\centering
\includegraphics[width=\linewidth]{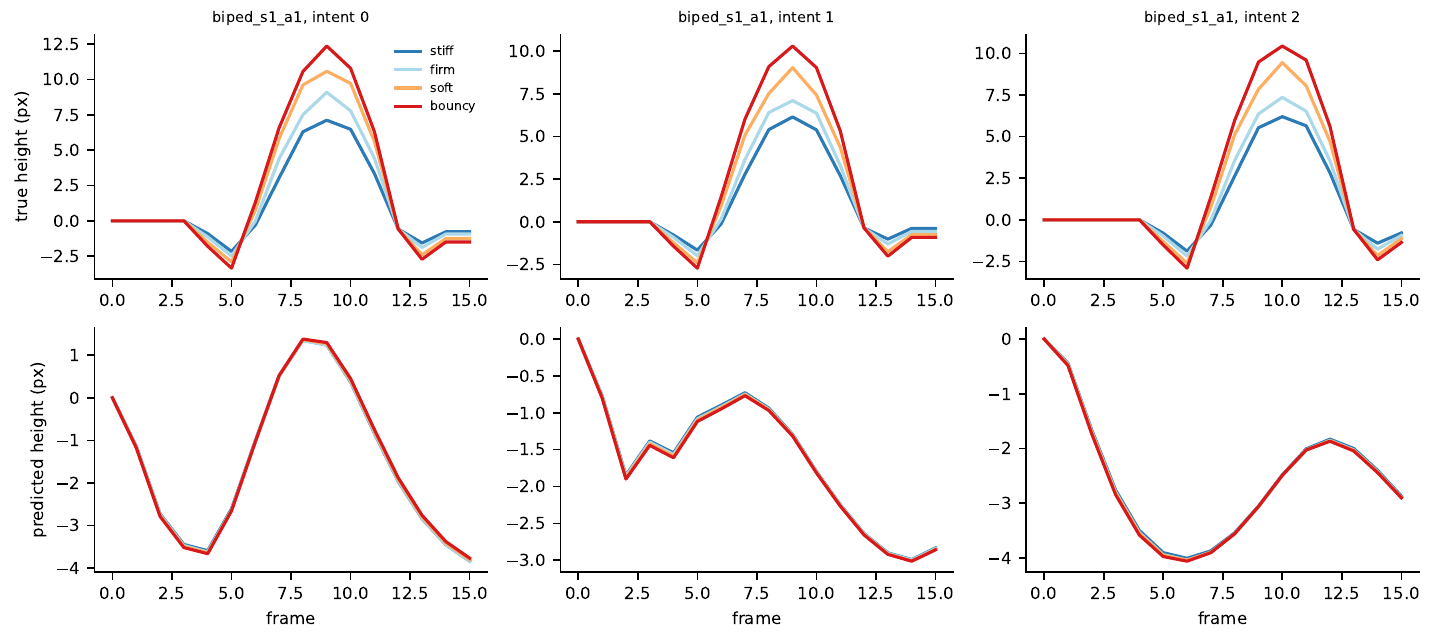}
\caption{Height of the four twins of one biped over one jump, for three intents. The top row shows the truth, and the bottom row shows the pixel world model, whose four curves overlap.}
\label{fig:exp3heights}
\end{figure}

\Cref{fig:exp3heights} shows this for one twin set. The true peak heights of the four twins differ by about 4 to 5 pixels, while the four predicted height curves lie on top of each other. We see two likely sources of this failure, although our experiments do not separate them. First, the decoder appears to average over the possible futures, so the four twins collapse into one blurred body. Second, the training objective itself appears to discard body information. A path head trained on the frozen body code of the pixel model reaches a twin matching of 0.48 with the plain code, which is lower than with the same encoder at its random initial weights (0.55, $p<0.001$). With the contrastive code it reaches 0.50, which is also no better than the random encoder ($p=0.084$). A likely reason is that the silhouette measures of the context clips already carry body information, and a pixel objective has no reason to keep it, because the average motion already overlaps every twin well.

\subsection{Body dynamics can be recovered from other actions}

\begin{figure}[H]
\centering
\includegraphics[width=\linewidth]{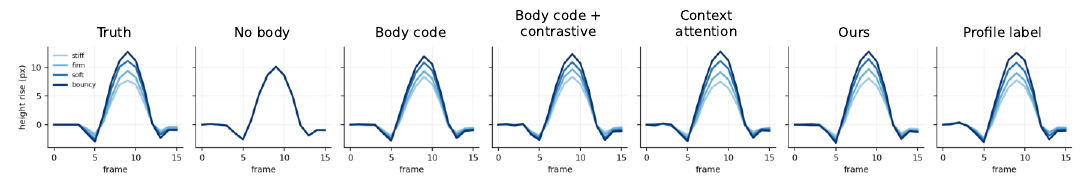}
\caption{Predicted height of the four twins of one biped over one jump. Without body information, all twins get the same path, while our model follows the true order and spread.}
\label{fig:paths}
\end{figure}

Predicting the motion path instead of pixels recovers most of the twin difference. \Cref{tab:paths} compares all path models on the same folds and seeds, and \cref{fig:paths} shows one twin set.

\begin{table}[H]
\centering
\caption{Path prediction on ToonTwins (means with 95\% bootstrap CIs over units). The swap column gives twin matching when each twin receives the context of another twin in its set, so a low value means that the prediction follows the context.}
\label{tab:paths}
\small
\begin{tabular}{lcccc}
\toprule
& \multicolumn{2}{c}{Twin matching (chance 0.25)} & Jump gain & Swap \\
\cmidrule(lr){2-3}
Model & New shapes & Unseen body plans & New shapes & New shapes \\
\midrule
No body & 0.250 & 0.250 & 0.00 & -- \\
Body code & 0.675 [0.630, 0.727] & \textbf{0.436} [0.385, 0.485] & 0.86 & 0.098 \\
Body code + contrastive & 0.672 [0.621, 0.728] & 0.389 [0.351, 0.424] & 0.82 & 0.106 \\
Context attention & 0.767 [0.699, 0.829] & 0.412 [0.357, 0.465] & 0.90 & 0.070 \\
\textbf{Ours} & \textbf{0.859} [0.820, 0.897] & 0.422 [0.375, 0.466] & \textbf{0.96} & 0.044 \\
\midrule
Profile label (reference) & 0.813 [0.787, 0.839] & 0.531 [0.470, 0.591] & 0.98 & 0.046 \\
\bottomrule
\end{tabular}
\end{table}

On new shapes, our model improves twin matching over the context attention baseline by 0.092 (95 percent CI 0.063 to 0.126, $d_z=1.32$, $p<0.001$) and over the body code by 0.184 ($d_z=3.73$, $p<0.001$). It also exceeds the model that receives the true profile label by 0.046 ($p=0.012$). This is because the context clips show how this exact shape moves, which the label does not. Swapping the contexts of twins drops twin matching to 0.044, below the chance level of 0.25, so the prediction follows the body information it receives. This test changes the context on purpose and does not prove that the model has identified the dynamics profile. Fine-tuning the model without body information on the context clips reaches only 0.281. The contrastive loss alone does not help the body code (0.672 against 0.675), so its gain on new shapes depends on the transformer (\cref{sec:ablation}). Every learned model ranks the four jump heights on new shapes with a Spearman correlation of 0.99 to 1.00.

\begin{figure}[H]
\centering
\includegraphics[width=0.8\linewidth]{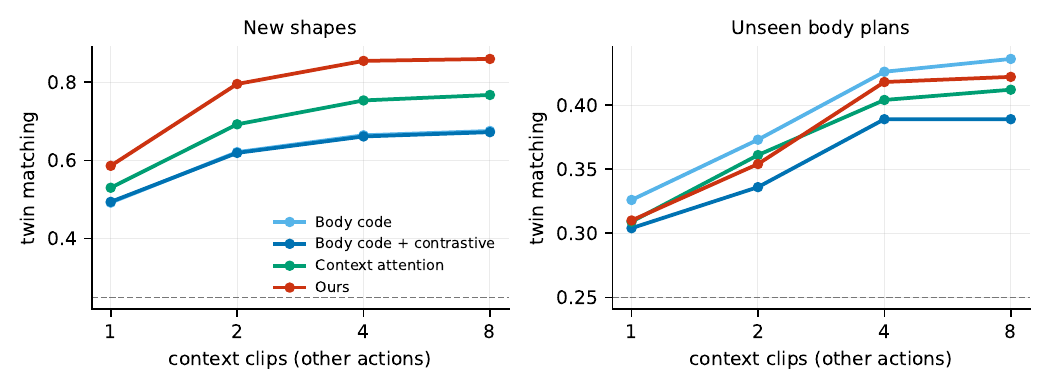}
\caption{Twin matching against the number of context clips. The dashed line marks chance.}
\label{fig:context}
\end{figure}

Two context clips already give most of the improvement (\cref{fig:context}). Our model reaches 0.586 with one clip and 0.795 with two clips, which is already above the context attention baseline with eight clips (0.767). With eight clips it reaches 0.859. A linear probe recovers the dynamics profile from our summary with 0.835 accuracy and the held-out action with 0.236 accuracy, close to the chance level of 0.20. The body code without the contrastive loss gives 0.655 and 0.760, so it stores which actions it has seen.

\begin{figure}[H]
\centering
\includegraphics[width=\linewidth]{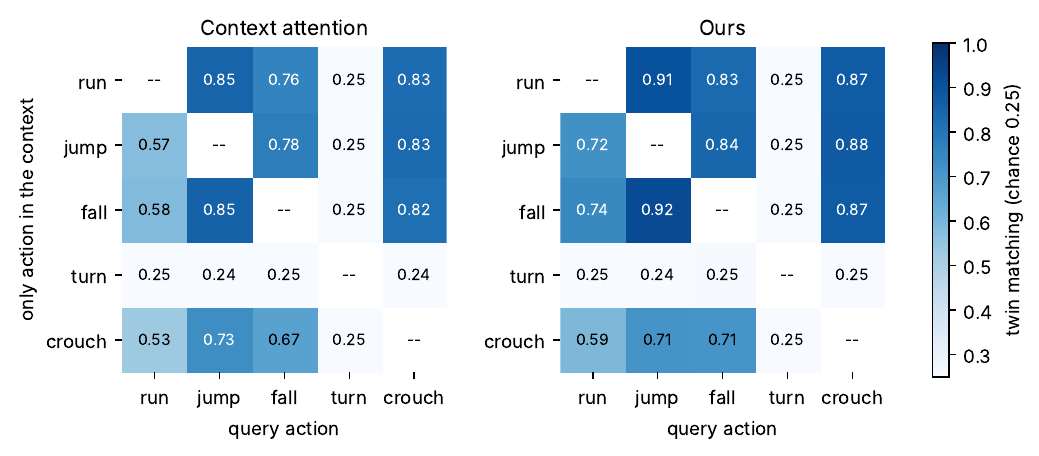}
\caption{Twin matching when the context holds only one action type (rows), for each query action (columns). Turning carries no body information.}
\label{fig:ctxinf}
\end{figure}

Not every action carries information about the body (\cref{fig:ctxinf}). A context of turns alone leaves every model at chance (0.24 to 0.25), because turning is the one action that does not depend on the dynamics profile. Running, falling and jumping are the most informative context actions, with 0.81 to 0.87 for our model averaged over the query actions other than turn. Crouching is the least informative (0.67). Our model is better than the context attention baseline for every informative context action.

\begin{figure}[H]
\centering
\includegraphics[width=1\linewidth]{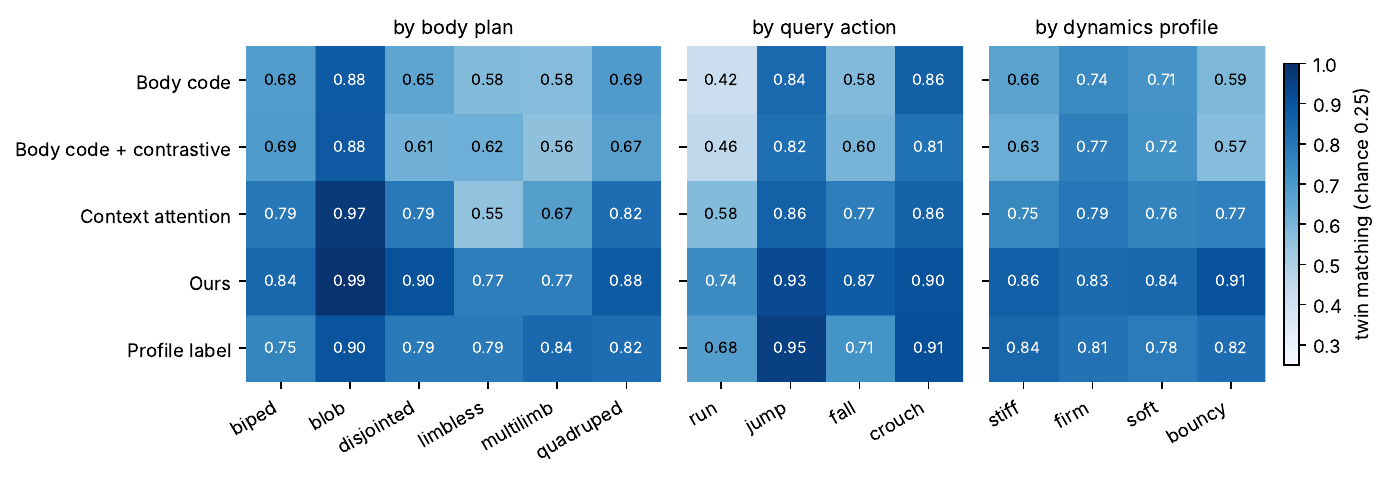}
\caption{Twin matching on new shapes by body plan, query action and dynamics profile (8 context clips, chance 0.25).}
\label{fig:breakdown}
\end{figure}

The lead of our model is largest for the extreme profiles. It matches bouncy twins in 0.906 of cases and stiff twins in 0.863, against 0.767 and 0.748 for the context attention baseline. \Cref{fig:breakdown} breaks twin matching down by body plan, query action and dynamics profile. Among the learned models, ours is best for every body plan, action and dynamics profile. Its largest gains over the context attention baseline are for limbless bodies (0.554 to 0.774), disjointed bodies (0.789 to 0.902), running (0.580 to 0.739) and falling (0.774 to 0.872). The profile label is better only for limbless and multilimb bodies and for jumping and crouching, where the motion depends on only one or two dynamics parameters, such as the jump gain or the crouch depth, so knowing the profile is enough.

\subsection{What each part of the model contributes}
\label{sec:ablation}

\Cref{tab:paths} does not include our model without the contrastive loss, so we retrained four models on the same folds and seeds to separate the parts of our model (\cref{tab:ablation}). We used 3 seeds for new shapes and for the animated animals and one seed for unseen body plans, so we treat the comparisons on unseen body plans as exploratory. The rerun reproduces the main results, with 0.857 against 0.859 for our model and 0.770 against 0.767 for the context attention baseline. The columns for the animated animals and unseen body plans are discussed in \cref{sec:real,sec:unseen}.

\begin{table}[H]
\centering
\caption{Controlled ablation with the same folds and seeds, from a separate rerun. TM is twin matching with 8 context clips (chance 0.25).}
\label{tab:ablation}
\small
\resizebox{\linewidth}{!}{%
\begin{tabular}{lcc cc cc cc}
\toprule
 & & & \multicolumn{2}{c}{New shapes} & \multicolumn{2}{c}{Unseen body plans} & \multicolumn{2}{c}{Animated animals} \\
\cmidrule(lr){4-5}\cmidrule(lr){6-7}\cmidrule(lr){8-9}
Model & Body token + query-free encoder & Contrastive sets & TM & Jump gain & TM & Jump gain & Reduction (\%) & Identification \\
\midrule
Context attention & -- & -- & 0.770 & 0.91 & \textbf{0.420} & \textbf{0.64} & 5.2 & 0.206 \\
Ours, contrastive loss off & \checkmark & -- & 0.806 & 0.93 & 0.418 & \textbf{0.64} & 3.2 & 0.153 \\
\textbf{Ours} & \checkmark & disjoint clips & \textbf{0.857} & \textbf{0.96} & 0.409 & 0.49 & 8.1 & 0.296 \\
Ours, action-disjoint sets & \checkmark & disjoint actions & 0.838 & 0.94 & 0.375 & 0.38 & \textbf{10.5} & \textbf{0.309} \\
\bottomrule
\end{tabular}}
\end{table}

On new shapes, giving the transformer a body token and a context encoder that does not see the query raises twin matching from 0.770 to 0.806 (difference 0.037, 95 percent CI 0.017 to 0.060, $d_z=0.77$, $p<0.001$). These two changes were made together, so their separate effects are not known. Adding the contrastive loss to the same architecture raises twin matching further to 0.857 (difference 0.050, CI 0.036 to 0.066, $d_z=1.48$, $p<0.001$). The contrastive loss gives more than half of the gain over the baseline, and this part cannot come from extra capacity, because the model without the loss has exactly the same parameters. Drawing the two contrastive sets from disjoint action types gives 0.838, which is lower than 0.857 ($p<0.001$) but still above the model without the loss ($p<0.001$), so the benefit of the loss does not depend on the two sets sharing action types.

\begin{figure}[H]
\centering
\includegraphics[width=1\linewidth]{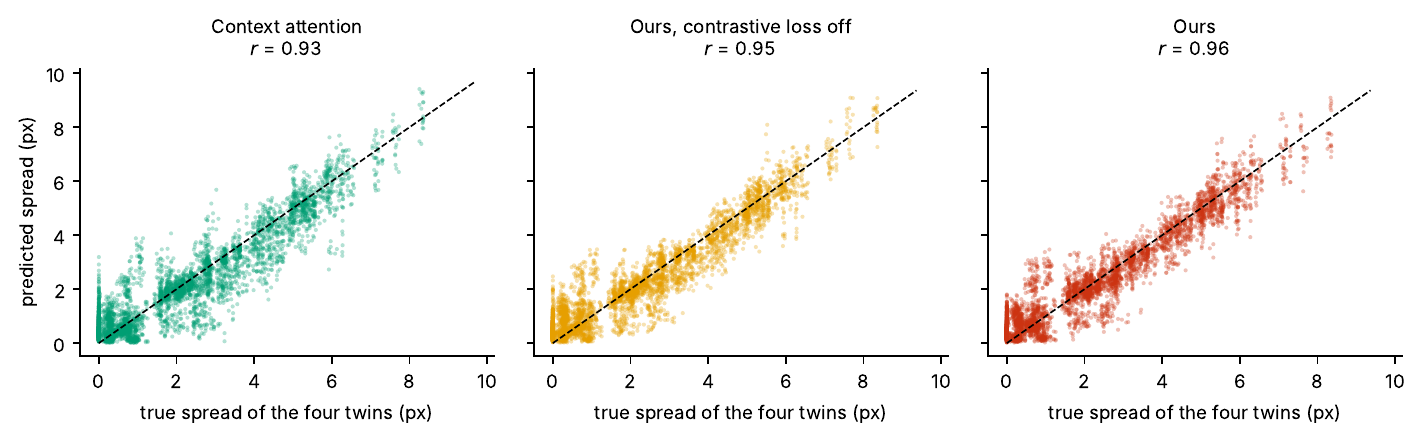}
\caption{Predicted against true spread of the four twins' peak heights for each twin set on new shapes. The dashed line marks perfect prediction.}
\label{fig:spread}
\end{figure}

\Cref{fig:spread} shows the same effect for each twin set. For several twin sets with a true spread of 2 to 3 pixels, the context attention baseline predicts a spread close to zero. Most of these cases disappear with the body token and the loss, and the correlation with the true spread rises from 0.93 to 0.95 and 0.96.

\subsection{The contrastive loss makes the summary more body-specific and less action-specific}
\label{sec:invariance}

The contrastive loss makes the summaries of one character agree across different context actions and separates look-alike twins, but it reduces rather than removes the information about the observed actions. A probe on the query action cannot show this, so we measured it directly. For each held-out character, we computed ten summaries, each from four clips of one pair of actions, and compared summaries from pairs that share no action (\cref{tab:invariance}).

\begin{table}[H]
\centering
\caption{Invariance of the body summary for held-out characters (95\% CIs over runs in brackets). The gap is the cosine for the same body minus the cosine for a twin, computed before rounding. Chance is 0.25 for twin retrieval and 0.10 for the action probe.}
\label{tab:invariance}
\small
\resizebox{\linewidth}{!}{%
\begin{tabular}{llcccccc}
\toprule
Split & Model & Cos, same body & Cos, twin & Gap & Retrieval & Action probe ($\downarrow$) & Profile probe \\
\midrule
New shapes & Silhouette features, no learning & -- & -- & -- & 0.329 [0.315, 0.343] & 0.999 [0.998, 1.000] & 0.506 [0.491, 0.524] \\
 & Ours, contrastive loss off & 0.987 & 0.833 & 0.154 [0.136, 0.173] & 0.792 [0.769, 0.820] & 0.626 [0.592, 0.661] & 0.651 [0.631, 0.668] \\
 & \textbf{Ours} & 0.813 & 0.248 & \textbf{0.564} [0.536, 0.597] & \textbf{0.873} [0.862, 0.887] & 0.139 [0.129, 0.150] & \textbf{0.799} [0.767, 0.837] \\
 & Ours, action-disjoint sets & 0.791 & 0.250 & 0.541 [0.506, 0.579] & 0.852 [0.829, 0.877] & \textbf{0.127} [0.119, 0.135] & 0.765 [0.722, 0.816] \\
\midrule
Unseen body plans & Ours, contrastive loss off & 0.959 & 0.834 & \textbf{0.126} [0.016, 0.314] & \textbf{0.450} [0.304, 0.622] & 0.545 [0.465, 0.656] & \textbf{0.353} [0.317, 0.390] \\
 & \textbf{Ours} & 0.214 & 0.103 & 0.110 [0.044, 0.173] & 0.369 [0.291, 0.443] & \textbf{0.104} [0.088, 0.122] & 0.326 [0.260, 0.423] \\
 & Ours, action-disjoint sets & 0.209 & 0.125 & 0.083 [0.000, 0.177] & 0.342 [0.258, 0.440] & 0.125 [0.103, 0.147] & 0.267 [0.196, 0.344] \\
\bottomrule
\end{tabular}}
\end{table}

Without the contrastive loss, the summaries of different characters stay close to each other. Two summaries of the same body have a cosine similarity of 0.987, but a body and its twin also reach 0.833. A linear probe recovers which pair of actions was observed in 0.626 of cases, where chance is 0.10. With the loss, two summaries of the same body from completely different actions stay similar (0.813), while twins drop to 0.248. The gap between the two grows from 0.154 to 0.564 ($d_z=6.7$, $p<0.001$ over 9 runs). The probe for the observed actions falls to 0.139 (95 percent CI 0.129 to 0.150). This is close to the chance level of 0.10 but above it, so the loss reduces this information without removing it.

\begin{figure}[H]
\centering
\includegraphics[width=0.8\linewidth]{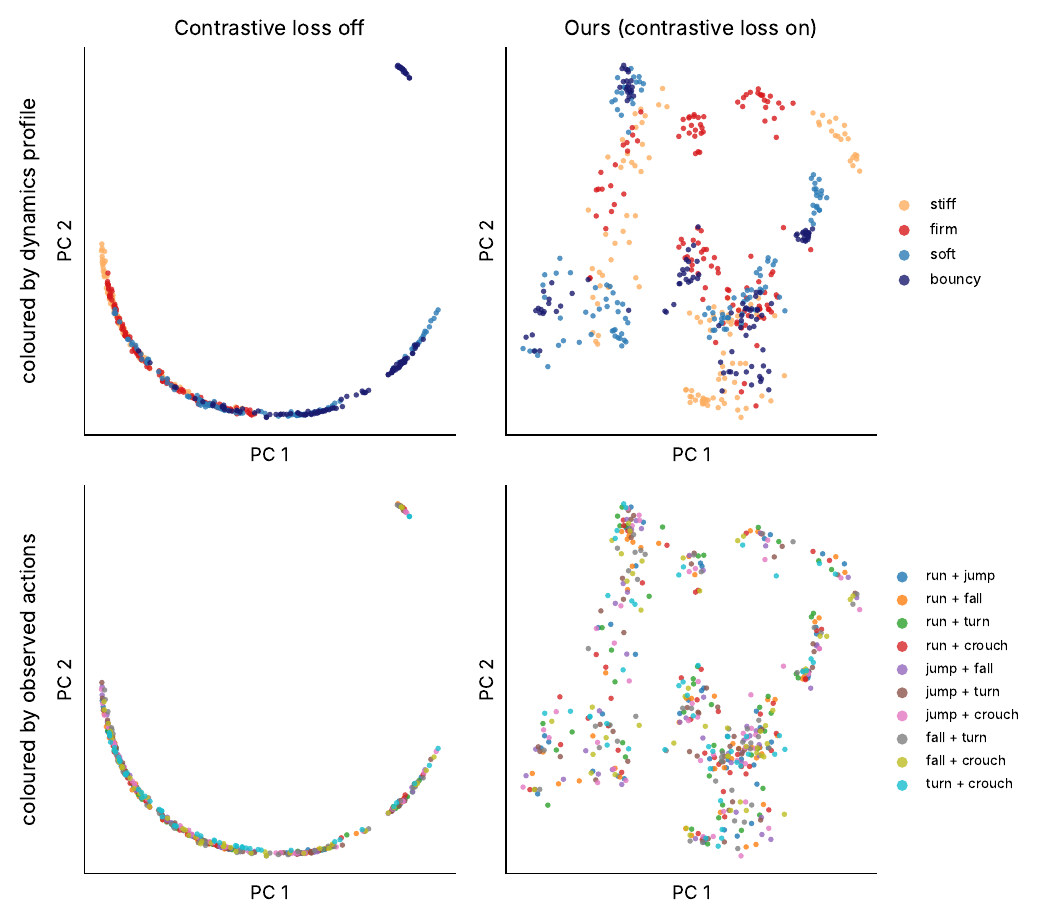}
\caption{Body summaries of held-out characters (first two principal components). Without the contrastive loss, they lie on one curve, and with it, each character forms its own cluster.}
\label{fig:clusters}
\end{figure}

The runs share test characters, so we also resampled over the 12 twin sets of one run (fold 0, seed 0). This gives the same result, with a gap of 0.608 (95 percent CI 0.516 to 0.694) with the loss and 0.187 (0.114 to 0.277) without it, and the gap is larger with the loss for all 12 twin sets. Across the nine runs, it ranges from 0.51 to 0.64 with the loss and from 0.11 to 0.21 without it. A twin is picked out among its four look-alikes from actions that share nothing in 0.873 of cases, against 0.792 without the loss. \Cref{fig:clusters} shows the same result.

We also ran the same probes on hand-made silhouette statistics with no learning, to check that the clips contain body information that can be recovered at all. For each clip, we took the maximum, minimum, mean and standard deviation of each of the nine measures and the mean absolute change between frames. We pooled these over the four clips of a context by their mean and maximum. A linear probe recovers the dynamics profile from these features in 0.506 of cases (95 percent CI 0.491 to 0.524), against 0.799 for our summary and a chance level of 0.25. However, the same features reveal the observed pair of actions in 0.999 of cases and pick out the correct twin in only 0.329 of cases (\cref{tab:invariance}), because they mix the body with the actions. In both cases, the feature scaler and the probe, a logistic regression with an L2 penalty ($C=1$), are fit on the training characters only, with no hyperparameter search.

\subsection{Path-guided rendering gives sharp frames that keep the difference}

\begin{figure}[H]
\centering
\includegraphics[width=0.62\linewidth]{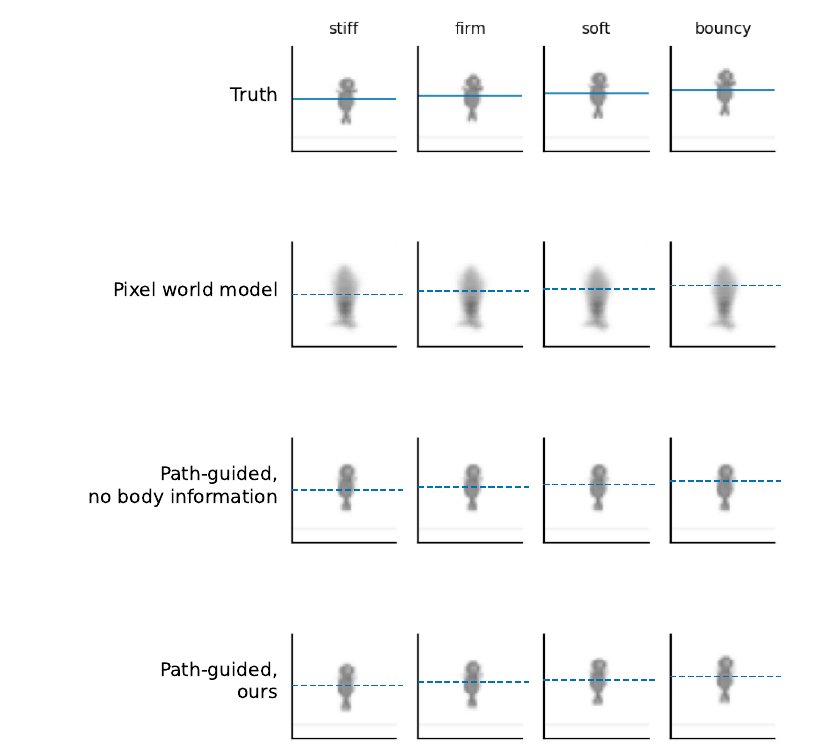}
\caption{Four look-alike twins at the peak of one jump, with lines at the true heights. The pixel model and the no-body renderer draw the same body for all four, while ours places each twin near its own height.}
\label{fig:frames}
\end{figure}

Drawing the frames along the predicted path keeps both the image quality and the twin difference (\cref{tab:render,fig:frames}).

\begin{table}[H]
\centering
\caption{Rendered frames on new shapes, compared with the truth at $48\times48$ pixels.}
\label{tab:render}
\small
\begin{tabular}{lcccc}
\toprule
Method & IoU & Sharpness & Twin matching & Jump gain \\
& & (1 = truth) & in images & in images \\
\midrule
Copy first frame & 0.502 & 0.97 & 0.250 & 0.00 \\
Pixel world model & 0.448 & 0.66 & 0.300 & 0.25 \\
No body, path-guided & 0.799 & 0.94 & 0.250 & 0.00 \\
Context attention, path-guided & 0.846 & 0.94 & 0.822 & 0.89 \\
\textbf{Ours, path-guided} & \textbf{0.861} & 0.94 & \textbf{0.904} & \textbf{0.95} \\
\midrule
True path, path-guided (renderer limit) & 0.888 & 0.95 & 0.991 & 1.00 \\
\bottomrule
\end{tabular}
\end{table}

Our rendered frames improve IoU over the pixel world model by 0.413 ($d_z=7.72$, $p<0.001$) and twin matching by 0.604 ($d_z=10.94$, $p<0.001$). Against the context attention baseline, twin matching in the images rises by 0.081 ($p<0.001$). The model without body information also renders accurate frames (IoU 0.799), but it gives all four twins the same motion (third row of \cref{fig:frames}). The gain in twin matching comes from the body information in the context and not from the renderer. The refiner adds 0.096 IoU and 0.168 twin matching over the warp alone. Running is the hardest action, because the warp moves the whole body and cannot swing the legs. Even so, our model matches running twins in 0.818 of cases.

\subsection{Animated animals}
\label{sec:real}

\begin{figure}[H]
\centering
\includegraphics[width=\linewidth]{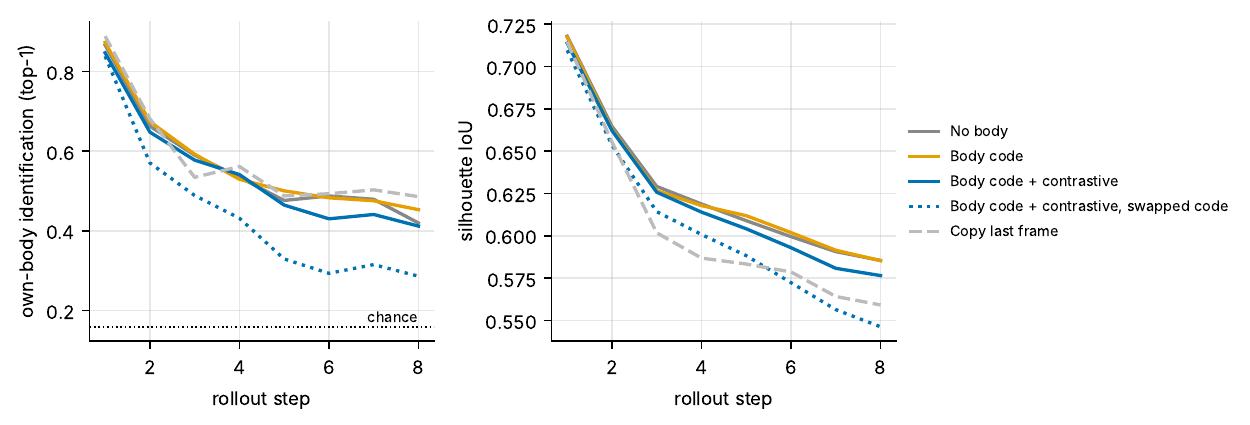}
\caption{Pixel world model on the animated animals. The left panel shows how often the rollout is closer to the animal's own future than to other animals' futures (chance 0.159), and the right panel shows the silhouette IoU.}
\label{fig:exp1curves}
\end{figure}

\begin{figure}[H]
\centering
\includegraphics[width=0.85\linewidth]{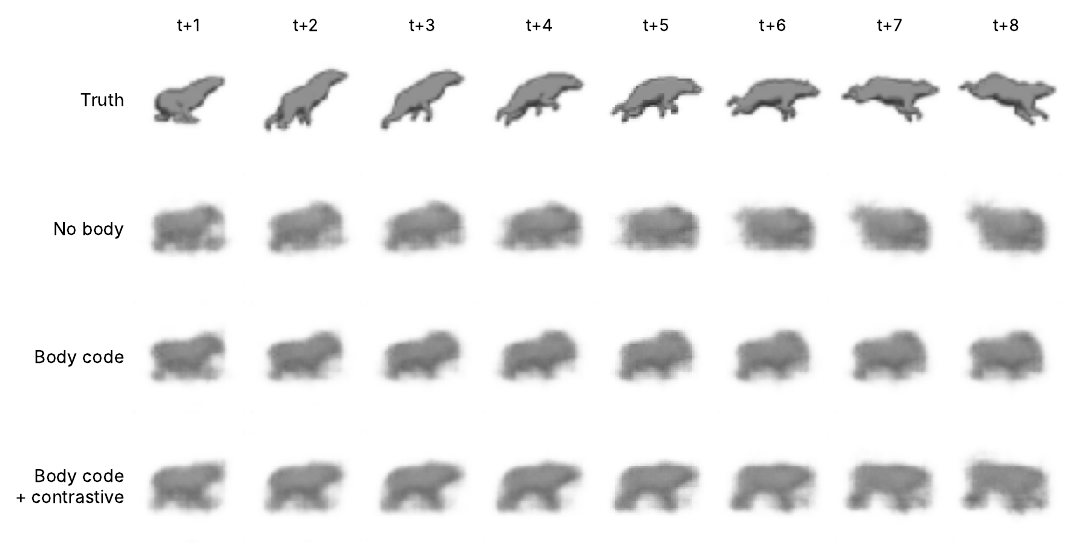}
\caption{Pixel world model rollouts for a held-out bear performing a jump absent from its context. The model draws a blurred average body.}
\label{fig:exp1rollout}
\end{figure}

To test whether the failure of the pixel world model on ToonTwins also holds for more complex bodies, we first trained it on these animals and asked it to roll out 8 frames of an action that is absent from the context, for held-out motion groups (\cref{fig:exp1curves,fig:exp1rollout}). We then measured how often the rollout is closer to the animal's own true future than to the futures of other held-out animals performing the same action (chance 0.159). This chance level differs slightly from the 0.157 of the path models because this test used an earlier split and a different set of candidate animals. The score was 0.565 without body information, 0.573 with a body code and 0.545 with the contrastive body code, and neither body code differs significantly from no body information ($p=0.45$ and $p=0.13$). Repeating the last observed frame reached 0.580. The model does use the contrastive code, because replacing it with another animal's code lowers the score by 0.101 ($p<0.001$), and a linear probe recovers the animal family from it with 0.565 accuracy (chance 0.20). The pixel model receives body information and reacts to it, but its rollouts do not improve and remain a blurred average body (\cref{fig:exp1rollout}), as on ToonTwins. This test used the first grouping with 24 groups, in which an animal could share its animations with a training animal. This can only help the pixel model, which still showed no gain, so we did not repeat it.

Our path model, in contrast, lowers the mean path error by 8.5 percent against the model without body information (95 percent CI 2.7 to 14.4), from 0.847 to 0.774 in standardized units ($p=0.004$). It identifies the animal from its own context in 0.324 of cases, where chance is 0.157 (\cref{tab:real}). The gain is uneven across queries. The error falls for 61 percent of them over all folds and seeds, and the percentage averaged over queries is only 5.0 (CI $-3.4$ to 12.0). Unlike on ToonTwins, more context lowers the error by less than 2 percent. With one, two or three other actions in the context, the error of our model is 0.787, 0.780 and 0.774, and the other models are similarly flat. A single other action of an animal already carries most of the body information that these models can use. Our model also has a lower error than the model that knows the animal family (0.810), although this difference is not significant ($p=0.13$).

\begin{table}[H]
\centering
\caption{Animated animals from DeformingThings4D (21 motion groups, 4 folds, 3 seeds).}
\label{tab:real}
\small
\begin{tabular}{lcc}
\toprule
Model & Error reduction vs.\ no body (\%) & Context identification (chance 0.157) \\
\midrule
Body code & 3.7 [$-$0.1, 7.3] & 0.274 \\
Body code + contrastive & 7.5 [3.4, 11.9] & \textbf{0.327} \\
Context attention & 4.8 [0.3, 9.3] & 0.193 \\
\textbf{Ours} & \textbf{8.5} [2.7, 14.4] & 0.324 \\
\midrule
Family label (reference) & 4.3 [$-$1.5, 10.5] & 0.390 \\
\bottomrule
\end{tabular}
\end{table}

\begin{figure}[H]
\centering
\includegraphics[width=0.78\linewidth]{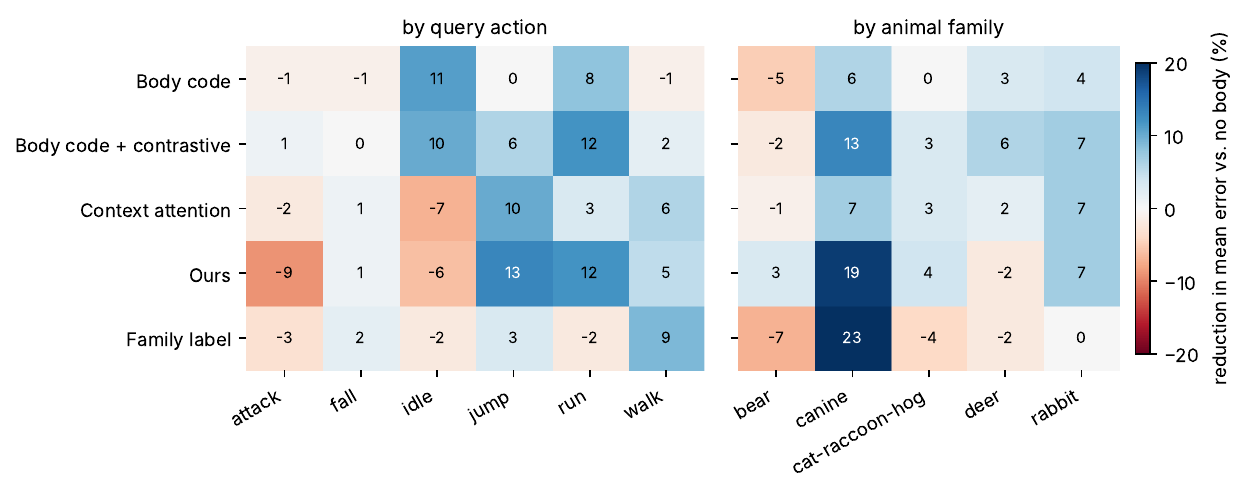}
\caption{Error reduction against the no-body model on the animated animals, by query action and animal family. Blue cells are better than no body, and red cells are worse.}
\label{fig:realbreak}
\end{figure}

Our model beats the context attention baseline on identification ($p=0.004$) and on the gap between another animal's context and its own ($p<0.001$), but its lower error is only borderline ($p=0.046$, Wilcoxon $p=0.16$). It matches the body code with the contrastive loss on error ($p=0.52$) and identification ($p=0.88$). The controlled rerun in \cref{tab:ablation}, which gives 8.1 percent for our model and 5.2 percent for the baseline, shows where the gain comes from. The body token and the query-free encoder do not help on these animals, and without the loss they give a slightly higher error than the baseline ($p=0.049$). Adding the contrastive loss lowers the error ($p=0.003$) and raises the identification of the animal from its own context from 0.153 to 0.296 ($p<0.001$). Drawing the two contrastive sets from disjoint action types gives 10.5 percent, slightly better than our model ($p=0.047$, Wilcoxon $p=0.13$). Most of the gain on these animals therefore comes from the contrastive loss and not from the transformer. \Cref{fig:realbreak} breaks the error reduction down by action. The largest gains of our model are for jumping (12.5 percent) and running (11.9 percent), while attacking and standing idle become worse (by 8.6 and 6.2 percent).

\begin{figure}[H]
\centering
\begin{subfigure}{\linewidth}\centering\includegraphics[width=0.8\linewidth]{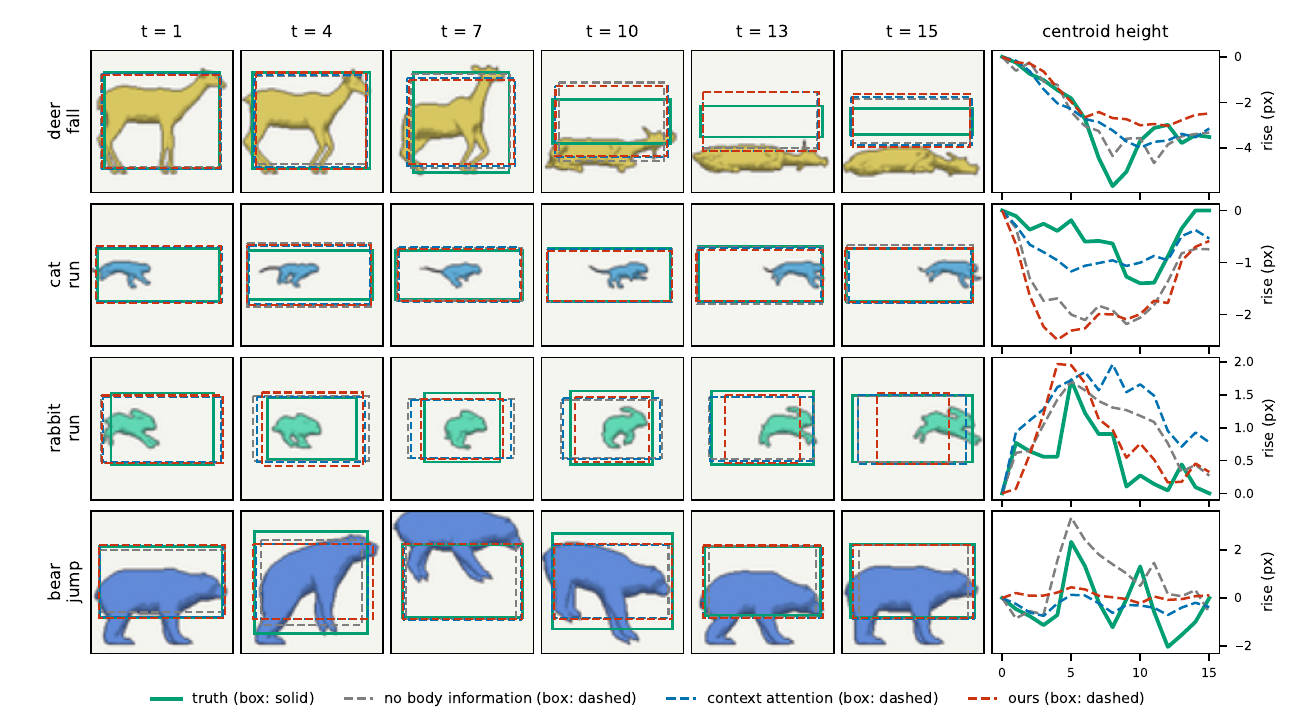}\caption{Typical cases (improvement closest to the median, from four animal families).}\end{subfigure}\\[4pt]
\begin{subfigure}{\linewidth}\centering\includegraphics[width=0.8\linewidth]{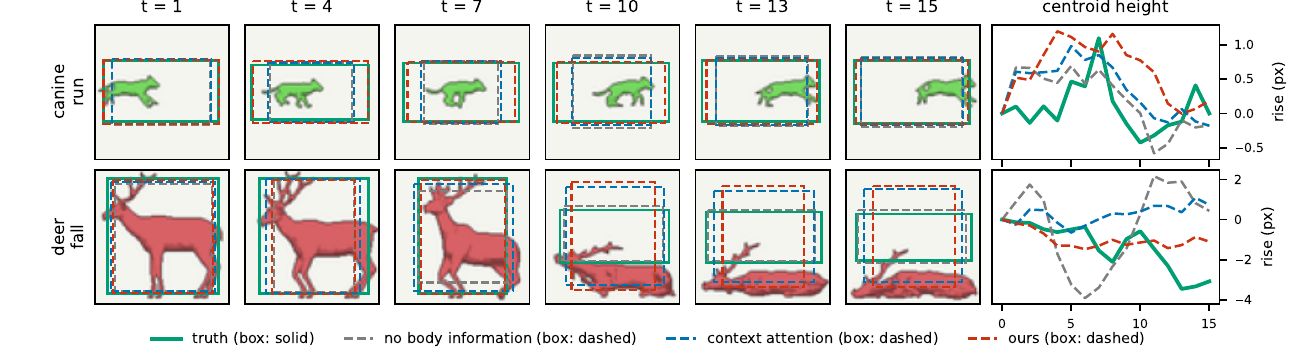}\caption{The best case (top) and the worst case (bottom).}\end{subfigure}
\caption{Predicted motion on held-out animals. The solid box is the true silhouette box, the dashed boxes are predictions from the same first frame, and the right column shows the rise of the centroid.}
\label{fig:realex}
\end{figure}

\Cref{fig:realex} shows actual predictions of the seed-0 models. We chose typical cases, whose improvement over the model without body information is closest to the median, together with the best and the worst case, so that the figure does not show only successes. Across all 169 held-out pairs of animal and action in this seed, the path error was 3.02\,px for our model, 3.08\,px for the context attention baseline and 3.15\,px without body information. Measured in pixels, our model improved on the model without body information in 63 percent of the cases of this seed. The gain was largest for canines (3.07 to 2.58\,px), while for deer our model was worse than the model without body information (3.35 against 3.21\,px). The boxes also show the main limit on these animals. Whole-body position and size are predicted well, but the differences between the models are small compared with the motion of the limbs, which a box does not describe.

\subsection{Unseen body plans remain hard}
\label{sec:unseen}

\begin{figure}[H]
\centering
\includegraphics[width=0.62\linewidth]{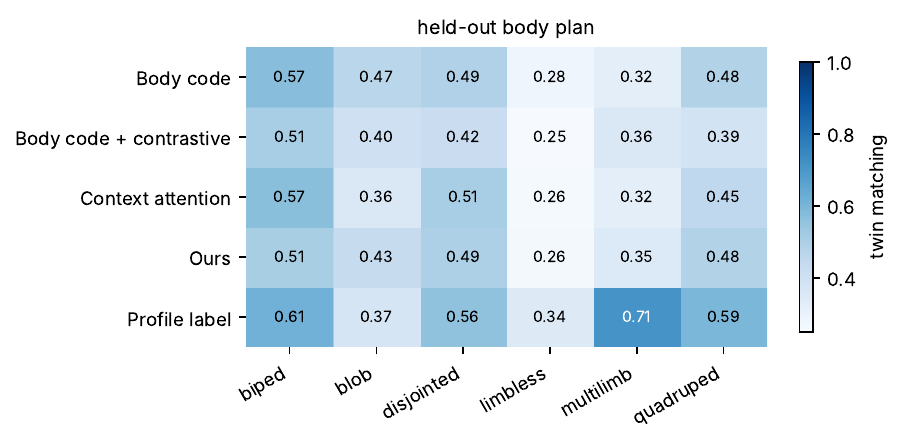}
\caption{Twin matching when the whole body plan is held out (columns), with 8 context clips (chance 0.25).}
\label{fig:lopoplans}
\end{figure}

When a whole body plan is held out, every learned model in the main comparison (\cref{tab:paths}) drops to a twin matching between 0.389 and 0.436, and the profile label reaches 0.531. Our model (0.422) is not different from the context attention baseline ($p=0.47$) or the body code ($p=0.20$). The limbless body plan stays close to chance for all learned models, at 0.25 to 0.28 (\cref{fig:lopoplans}). For the multilimb body plan, the profile label reaches 0.71, while every learned model stays at or below 0.36. The context clips of known body plans do not show the models how a new kind of body responds to its dynamics profile. In rendered frames, our model also ranks the jump heights less reliably on unseen body plans, and it identifies the highest jumper in 0.59 of cases against 0.78 for the context attention baseline.

The ablation suggests a reason, although it used one seed on this split and is exploratory (\cref{tab:ablation,tab:invariance}). On unseen body plans, the contrastive loss does not improve twin matching (0.409 and 0.375 against 0.418 without it) and lowers the jump gain from 0.64 to 0.49 and 0.38 ($p<0.001$). Two summaries of the same new body reach a cosine similarity of only 0.21, so the consistency learned on known body plans does not carry over to a new one. One possible explanation is that the loss learns distinctions within the body plans seen in training, rather than a summary that applies to unfamiliar structures. Testing this needs a wider range of body plans.

\section{Discussion and Limitations}

\subsection{Pixel metrics and body dynamics}

Pixel error and IoU should not be used alone to judge whether a world model adapts to a character. In our tests, the pixel world model had an IoU only 0.05 below copying the first frame, while it recovered at most 25 percent of the twin difference. A test with look-alike bodies shows this gap directly and needs only a procedural generator. The two models we compared also differ in their output and their training objective. The pixel world model predicts frames through a latent state, while our model predicts a silhouette path that a separate renderer draws. Our results show a limitation of the pixel model we tested, not that path prediction is better than pixel-based world modeling in general. Within our experiments, a simple warp and a small refiner draw sharp frames once the path is predicted. With the true paths, the renderer reaches an IoU of 0.888, so most of the remaining error comes from the predicted path and not from drawing.

\subsection{The role of the contrastive loss}

The contrastive loss controls what the body summary stores. With the same architecture and parameters, it raised twin matching on new shapes from 0.806 to 0.857 and made the summary of one body agree across actions that share nothing (\cref{sec:invariance}). On the animated animals, it was the only part of our model that helped. However, it lowered the jump gain on unseen body plans in our single-seed comparison, so the weight $\lambda$ of the loss may need a different value when the test contains new body plans.

\subsection{Known and new body plans}

For new shapes of known body plans, the model predicts body-specific motion well, and its summary of one body stays consistent across actions that share nothing. For body plans never seen in training, both prediction and consistency drop sharply (\cref{sec:unseen}). What the model learns is therefore bounded by the variety of body structures seen in training. Widening that variety, or describing a body by its parts so that new bodies can be built from known parts, is one possible next step.

\subsection{Limitations}

This study intentionally described each body by its whole silhouette, to isolate body-specific dynamics as the only experimental variable. Consequently, the renderer cannot move limbs independently, and running is the weakest action in rendered frames (twin matching 0.818). The model also does not transfer to unseen body plans, where the limbless body plan stays near chance. Both problems need a description of the parts of a body and how they move together.

ToonTwins uses simple 2D characters at $64\times 64$ pixels, and its dynamics profiles follow fixed rules in the generator. We did not test parameter values outside the four profiles used in training. Each profile also changes all eight parameters together, so our tests do not show which parameters the model uses to tell twins apart. The animal data contains only 32 animals in 21 motion groups, which limits the precision of these results, for example a 95 percent CI of 2.7 to 14.4 percent for the error reduction. The animals are also animations made by artists, so the gain shows an improvement on this subset and does not show transfer to natural animal video.

We did not evaluate large pretrained video generation models, since adapting them to condition on a few clips of a specific character would require substantial changes. We predict only 15 future frames, so the accumulation of error over longer rollouts is not tested.

\section{Conclusion}

The latent pixel world model we tested kept at most 25 percent of the difference between look-alike bodies. A model that encodes the character's other actions and predicts the motion path separated twins of new shapes in 85.9 percent of cases. Drawing frames along the predicted path kept this difference in the rendered frames (twin matching 0.904). More than half of the gain over a transformer baseline came from the contrastive loss, which made the summaries of one body agree across actions that share nothing. On 32 animated animals, the mean path error fell by 8.5 percent, and a simpler body code with the same loss gave a similar gain. These results suggest three points.
\begin{itemize}
\item World models for characters should be tested on look-alike characters that move differently, because pixel metrics do not show whether body dynamics are used.
\item In our experiments, carrying body information through a predicted motion path kept it in the rendered frames, whereas the tested pixel model lost it.
\item The body summary should be trained to depend as little as possible on which actions were observed.
\end{itemize}
The method does not transfer to body plans absent from training, and this transfer needs a wider range of body structures.

\appendix
\refstepcounter{section}
\label{app:params}
\section*{Appendix~\thesection\quad ToonTwins Parameters}
\addcontentsline{toc}{section}{\texorpdfstring{Appendix~\thesection\quad ToonTwins Parameters}{Appendix A ToonTwins Parameters}}

\Cref{tab:defaults} gives the default dynamics parameters of each body plan, and \cref{tab:profiles} gives the multipliers of each dynamics profile. The value used for a character is the default of its body plan times the multiplier of its profile. Heights are in pixels of the $128\times 128$ drawing canvas, which is downsampled to $64\times 64$ pixels, and the turn rate does not depend on the profile.

\begin{table}[H]
\centering
\caption{Default dynamics parameters of the six body plans.}
\label{tab:defaults}
\small
\begin{tabular}{lccccccccc}
\toprule
Body plan & Jump & Bounce & Squash & Bob & Gait & Run speed & Crouch & Fall & Turn rate\\
\midrule
Biped      & 28 & 0.05 & 0.5 & 1.5 & 0.80 & 3.0 & 0.45 & 1.4 & 1.4\\
Blob       & 40 & 0.35 & 1.6 & 6.0 & 0.60 & 2.2 & 0.70 & 1.3 & 0.8\\
Disjointed & 30 & 0.20 & 0.5 & 2.0 & 0.85 & 3.3 & 0.50 & 1.0 & 1.2\\
Limbless   & 18 & 0.02 & 0.8 & 0.0 & 1.10 & 2.6 & 0.60 & 1.5 & 0.7\\
Multilimb  & 22 & 0.15 & 0.4 & 0.6 & 1.50 & 3.6 & 0.35 & 1.1 & 1.6\\
Quadruped  & 34 & 0.10 & 0.6 & 1.2 & 0.95 & 4.2 & 0.55 & 1.6 & 1.0\\
\bottomrule
\end{tabular}
\end{table}

\begin{table}[H]
\centering
\caption{Multipliers that each dynamics profile applies to the default parameters of a body plan.}
\label{tab:profiles}
\small
\begin{tabular}{lcccccccc}
\toprule
Profile & Jump & Bounce & Squash & Bob & Gait & Run speed & Crouch & Fall\\
\midrule
Stiff  & 0.78 & 0.35 & 0.35 & 0.55 & 0.88 & 0.92 & 0.80 & 1.15\\
Firm   & 0.92 & 0.75 & 0.75 & 0.85 & 0.96 & 0.97 & 0.93 & 1.05\\
Soft   & 1.08 & 1.50 & 1.45 & 1.30 & 1.04 & 1.00 & 1.08 & 0.95\\
Bouncy & 1.22 & 2.30 & 2.00 & 1.75 & 1.12 & 1.03 & 1.20 & 0.86\\
\bottomrule
\end{tabular}
\end{table}

\end{document}